\documentclass{article}

\usepackage[main,final]{neurips_2026}

\usepackage[utf8]{inputenc} 
\usepackage[T1]{fontenc}    
\usepackage{hyperref}       
\usepackage{url}            
\usepackage{booktabs}       
\usepackage{amsfonts}       
\usepackage{nicefrac}       
\usepackage{microtype}      
\usepackage{xcolor}         
\usepackage{graphicx}       
\usepackage{amsmath}        
\usepackage{amssymb}        
\usepackage{caption}
\usepackage{multirow} 
\usepackage{wrapfig}

\usepackage{booktabs, colortbl}

\definecolor{headerblue}{RGB}{220, 230, 242}
\definecolor{groupgray}{RGB}{245, 245, 245}

\title{Paleoinspired Vision: From Exploring Colour Vision Evolution to Inspiring Camera Design}

\author{%
  \textbf{Yijie Lu}$^{1}$ \quad \textbf{Zhimin Zong}$^{2}$ \quad \textbf{Junjie Zhang}$^{3}$\\
  \textbf{Shenghan Su}$^{4}$ \quad \textbf{Zeyun Zhao}$^{5}$ \quad \textbf{Ziteng Cui}$^{4,6}$ \quad \textbf{Yan Pu}$^{7}$\\
  \textbf{Jing Lu}$^{7}$ \quad \textbf{Daisuke Kojima}$^{6}$ \quad \textbf{Lin Gu}$^{8}$\thanks{Corresponding author: \texttt{lin@tohoku.ac.jp}} \quad \textbf{Ruogu Fang}$^{5}$\\[3pt]
  $^{1}$Peking University \quad $^{2}$University of Florida \quad $^{3}$The University of Hong Kong\\
  $^{4}$The Hong Kong University of Science and Technology (Guangzhou)\\
  $^{5}$Vanderbilt University \quad $^{6}$The University of Tokyo\\
  $^{7}$Key Laboratory of Vertebrate Evolution and Human Origins, Institute of Vertebrate\\
  Paleontology and Paleoanthropology, Chinese Academy of Sciences\\
  $^{8}$Tohoku University
}

\begin{document}

\maketitle

\begin{abstract}
The evolution of colour vision is captivating, as it reveals the adaptive strategies of extinct species while simultaneously inspiring innovations in modern imaging technology. In this study, we present a simplified model of visual transduction in the retina, introducing a novel opsin layer. We quantify evolutionary pressures by measuring machine vision recognition accuracy on colour images shaped by specific opsins. Building on this, we develop an evolutionary conservation optimisation algorithm to reconstruct the spectral sensitivity of opsins, enabling mutation-driven adaptations to more effectively spot fruits or predators. This model condenses millions of years of evolution into seconds using GPU, providing an experimental framework to test long-standing hypotheses in evolutionary biology, such as vision of early mammals, primate trichromacy from gene duplication, retention of colour blindness, blue-shift of fish rod and multiple rod opsins with bioluminescence.
Beyond validating existing paleontological hypotheses, our model further enables speculative explorations of hypothetical species, such as organisms with visual systems adapted to Martian conditions. These constitute our contributions to AI for life science and neuroscience. Besides, the principles derived from these biological studies can translate into practical engineering solutions. In practice, our findings suggest a minimalist yet effective approach to task-specific camera filter design, optimising the spectral response function to meet application-driven demands. 
Code repository: \url{https://github.com/luyijie-research/PaleVision}.
\end{abstract}

\section{Introduction}
\label{sec:intro}

\textit{To suppose that the eye with all its inimitable contrivances, could have been formed by natural selection, seems, I freely confess, absurd in the highest possible degree.}\\
\rightline{- Charles Darwin   ``The Origin of Species'' }

\begin{figure*}
  \centering
  \includegraphics[width=\linewidth]{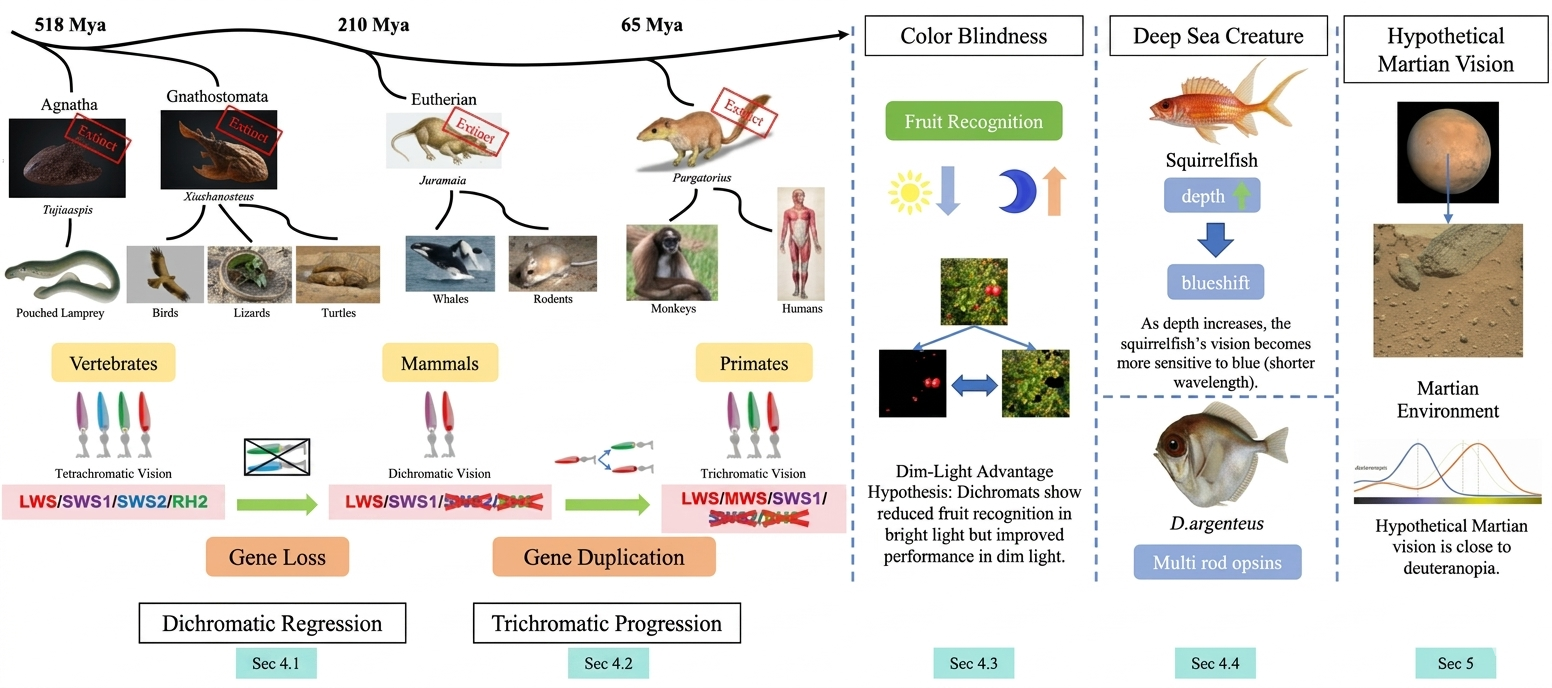}
  \caption{Our framework reconstructs the 518-million-year evolutionary trajectory of color vision across the vertebrate-to-primate lineage and evaluates corresponding explanatory hypotheses. This includes the transition from ancestral tetrachromacy through mammalian gene loss (dichromatic regression) to primate gene duplication (trichromatic progression). Furthermore, the framework quantifies the dim-light advantage hypothesis, deep-sea opsin blueshifts, and predictive visual adaptations and sensor designs for novel environments, such as the Martian landscape.}
  \label{fig:verte evolution}
\end{figure*}

Darwin argued that natural selection producing the eye 'can hardly be considered real.' However, modern biology has revealed eyes evolved not just once, but at least 40 different times \cite{Schwab2018}. Colour vision, in particular, reflects a tightly coupled interaction between environmental selection pressures and biological constraints, and continues to underpin modern camera design. As shown in Fig.\ref{fig:verte evolution}, early vertebrates exhibited tetrachromatic vision (4 colour vision) around 518 million years ago \cite{oldestverte, nocturnalbottleneck}. Early  mammals like Juramaia \cite{Luo2011}, which are only active in evening to avoid predation, then lost two colour in 210 million years ago, widely attributed attributed to this nocturnal lifestyles. Then, after  an asteroid struct earth  and wiped out dinosaurs approximately 65 million years ago, our primates ancestors returned to day life and  regained the third colour vision\cite{nocturnalbottleneck, evolutionofgeneduplication}. 
\enlargethispage{-3\baselineskip}
While prior palaeontological studies link evolutionary transitions to selection pressures, direct evidence is limited, and the timescales involved make these processes difficult to reproduce or quantitatively validate in controlled settings. Motivated by recent advances in reconstructing  evolutionary trajectories of both extant and extinct specie through controlled robotic experimental platforms, we propose a computational framework  called \textit{evolutionary conservation optimization} to quantitatively analyse enduring palaeontological hypotheses and translate the resulting principles into practical camera design solutions.

\begin{figure*}
  \centering
  \includegraphics[width=\linewidth]{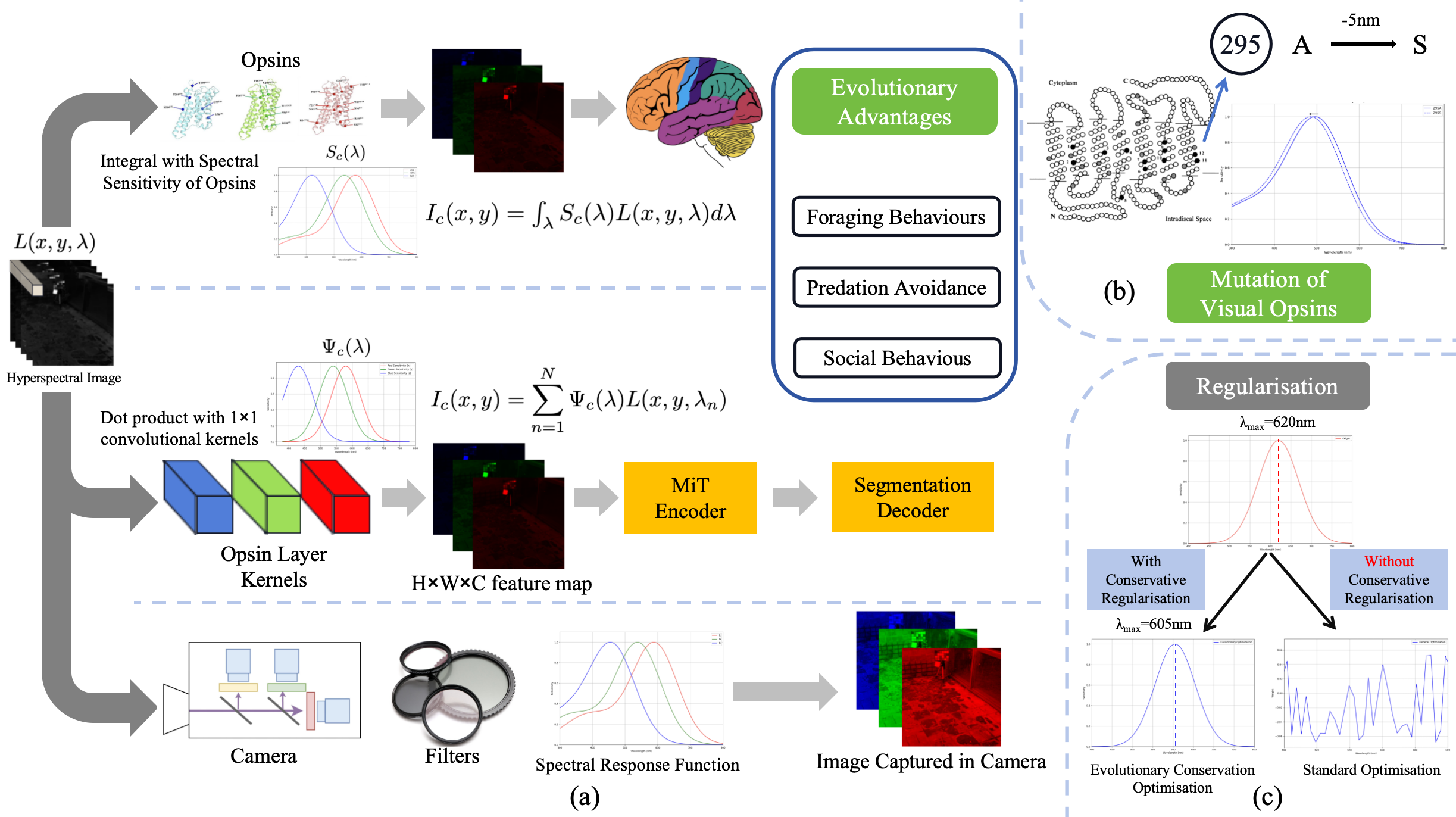}
  \caption{Computational framework for Evolutionary Conservation Optimization. The pipeline translates ecological pressures into semantic segmentation tasks, utilizing an opsin layer  to bridge low-level spectral perception with high-level decision-making. By incorporating conservative regularization—parameterizing kernels as Gaussian functions with constrained $\lambda_{max}$ shifts, the framework simulates the evolution and derive task-adaptive spectral response functions for specialized camera designs in Martian and medical applications.}
  \label{fig:main}
\end{figure*}

Our framework, shown in Fig. \ref{fig:main}, models visual evolution as a regularised optimisation problem over spectral representations. In Fig. \ref{fig:main}.a, we represent visual input as wavelength-dependent light signals and introduce an opsin-inspired layer to simulate colour perception that performs spectral filtering through learnable sensitivity functions. Each opsin is parameterized as a smooth, Gaussian-like response with a constrained peak wavelength  $\lambda_{max}$, ensuring biological plausibility\cite{GOVARDOVSKII_FYHRQUIST_REUTER_KUZMIN_DONNER_2000, LOEW19941427, Schott2022}. The filtered responses are processed by a vision backbone (MiT-Encoder) to perform downstream tasks such as semantic segmentation, which we use as a proxy for evolutionary pressure, capturing survival-relevant behaviours such as foraging and predator avoidance.

We simulate evolution by optimizing the spectral parameters based on task performance. Small changes in $\lambda_{max}$ represent mutation-like updates (Fig. \ref{fig:main}.b), while the task objective defines the fitness landscape. To keep the evolution realistic, we apply conservative regularisation that encourages smooth and interpretable spectral changes (Fig. \ref{fig:main}.c). This allows us to efficiently study how visual systems adapt to different environments in history. Beyond palaeontological analysis, we use the same framework to design cameras by changing the task objective in Fig. \ref{fig:main}.a. By optimizing spectral parameters for application-specific environments and tasks, we obtain task-adaptive sensitivity functions that can be directly translated into physical optical filters.

Using  this framework, we reconstruct key transitions in the evolution of colour vision, including the regression from tetrachromacy to dichromacy in early mammals and the re-emergence of trichromacy in primates. We further quantitatively analyse benefits of color deficiency under low-light conditions, adaptations in aquatic environments, such as the blue-shift of rod opsins with increasing depth and the emergence of multiple rod opsins in bioluminescent conditions. Together, these represent our contributions to the intersection of artificial intelligence with life science and neuroscience.

Beyond palaeontological analysis, the optimized spectral responses can be directly mapped to physical optical filters, enabling the design of task-adaptive camera system by optimizing spectral parameters for application-specific environments and tasks. We demonstrate this with proof-of-concept designs for 1) \textbf{Mars exploration}: By improving the segmentation of shadows and rocks on Mars, the designed filters show potential for cameras used in autonomous vehicle navigation in the Martian environment. 2) \textbf{Cancer detection}: Our design also demonstrates potential for medical-specific cameras capable of detecting abnormal tissue patterns associated with cancer. The evolutionarily grounded design paradigm is manufacturing-friendly, requiring only minor, easily implementable adjustments per iteration for rapid updates \cite{manufacturing}. This aligns with minimalist camera design principles \cite{Minimalist_ECCV24}. 

In summary, we present three main contributions: 
\begin{itemize}
    \item  A computational framework combining opsin layers with evolutionary conservation optimization to model opsin evolution trajectories.
    \item An experimental platform for reconstructing and quantifying colour vision transitions in mammals, primates and deep-sea species.
    \item A minimalist, task-adaptive camera design methodology optimizing spectral response functions for Martian exploration and cancer detection applications.
\end{itemize}
\section{Related Works}
\label{sec:related works}
\subsection{Relationship between Evolution Strategies and Gradient Descent}
\vspace{-2mm}
Recent research increasingly utilizes deep learning to simulate biological evolution \cite{https://doi.org/10.1111/2041-210X.13901}. Specifically, researchers have explored the intrinsic relationship between Evolution Strategies (ES) and gradient-based optimization \cite{ESSA}. While these paradigms are often considered fundamentally different, recent analytical work has demonstrated a formal correspondence between them; for instance, training neural networks via conditioned stochastic mutation or neuroevolution is equivalent, in the limit of small mutations, to gradient descent on the loss function in the presence of Gaussian white noise \cite{Whitelam2021}. This mathematical bridge suggests that gradient descent can be viewed as a simulator of mutation and selection pressure. Such a perspective aligns with findings that ES derive learning signals by stochastically perturbing parameters to perform randomized finite-difference estimation \cite{728207, Zhang2017OnTR}. To further integrate these fields, frameworks like Evolutionary Stochastic Gradient Descent (ESGD) combine SGD with gradient-free evolutionary steps to improve population fitness \cite{Cui2018ESGD}. Similarly, the Gradient Evolution (GE) algorithm incorporates the Newton-Raphson method into evolutionary operators to maintain diversity and avoid local optima \cite{KUO2015246}. These hybrid approaches complement theoretical investigations suggesting that natural selection may implicitly approximate natural gradient descent on statistical manifolds \cite{DaCosta2020NaturalSF}. Beyond weight optimization, evolutionary principles can synthesize entire network architectures, mimicking heredity to produce highly efficient models \cite{Shafiee2018}. Collectively, these works reinforce the hypothesis that gradient descent is not merely an artificial optimizer but a process that can naturally emerge from evolutionary dynamics \cite{11043088}, justifying its use for simulating complex evolution via backpropagation in our framework.


\subsection{Camera Design}

Traditional RGB cameras mimic human colour perception \cite{6475015} through multiple processing steps \cite{karaimer_brown_ECCV_2016}, but their spectral response functions are not optimized for specific imaging tasks \cite{nie2018deeply}. Empirical studies have modeled camera color pipelines to link output to scene attributes \cite{empiricalvision}, others have explored the potential of standard sensors \cite{ringlight} and joint sampling-reconstruction designs—using panchromatic-dominant patterns to maximize light intake and sharpness while suppressing aliasing \cite{rethinking}. Deep optics, which integrates optical design with deep learning \cite{PMID:33268862}, has been utilized to create higher-quality imaging systems \cite{cote2023differentiable, Li:21, Sun2021DiffLens, Tseng2021DeepCompoundOptics, Tseng2021NeuralNanoOptics, splitaperturecameras, 2023nanoarray, tian2025endomambaefficientfoundationmodel}. Pioneered by FlexISP \cite{flexisp}, this paradigm enables end-to-end optimization of the entire imaging chain \cite{differentiable, Split-Aperture}. Pre-sensor computing further accelerates these computations via optical neural networks (ONNs). Shi et al. \cite{shi2022loen} introduced a miniaturized lensless ONN, though it was limited by shallow layers. To overcome this, a multilayer ONN (MONN) was proposed \cite{doi:10.1126/sciadv.ado8516}, incorporating quantum dot films for non-linear activation \cite{miscuglio2018all} to enhance machine vision tasks.
Klotz and Nayar introduced minimalist vision with freeform pixels \cite{Minimalist_ECCV24}, utilizing arbitrarily shaped pixels to capture only task-essential information for privacy-preserving, self-sustaining sensing. Complementing these architectural shifts, the Tianmouc vision chip \cite{visionchip2024} demonstrates the power of co-designing optics and computation, achieving 10,000 fps and 90\% bandwidth reduction through dual-path processing for robust perception.

\section{Proposed Approach}
\label{sec:proposed approach}
\subsection{Overview}
Fig.\ref{fig:main} illustrates colour vision in vertebrates is mediated by opsins in the retina \cite{humaneye}, each with a distinct spectral sensitivity function, $S_c(\lambda)$, quantifying light detection efficiency across wavelengths ($\lambda$) for each specific opsin channel $c$. Given the spectral signal at position $(x,y)$, represented as $L(x,y,\lambda)$, the perceived intensity for each opsin is the integral of its spectral sensitivity over wavelength range, as shown by equation in Fig.\ref{fig:main}, generating a distinct signal channel for brain. We simplify this visual transduction using an \textit{opsin layer} with $c$ convolutional kernels, where each kernel's weight  $\Psi_c(\lambda)$ represents the spectral sensitivity function. The opsin layer performs a $1 \times 1$ convolution, which is equivalent to integration above\cite{nie2018deeply}, transforming the $H \times W \times N$  hyperspectral images (HSI) into a size $H \times W \times C$  feature map (where $C = 3$ for trichromatic vision). HSI, collected by hyperspectral cameras\cite{Li2024, 9878871, Bian2024}, capture the spectral per pixel, where $N$ represents the number of spectral bands (e.g. 400nm). The $C$ channel maps are then passed to  Mix Transformers (MiT) encoder for the recognition task.

\subsection{Opsin Layer}
\label{sec:response embedding}
The spectral sensitivity of opsins is commonly approximated using Gaussian functions, with a maximum sensitive wavelength defining their response to light \cite{GOVARDOVSKII_FYHRQUIST_REUTER_KUZMIN_DONNER_2000, LOEW19941427, Schott2022}. Therefore, we proposed a simple yet efficient computational model, namely opsin layer to simulate the spectral sensitivity function shown in Fig.\ref{fig:main}. Since the integration of spectral and spectral sensitivity function of opsins equvalents to dot product between entries of the convolution kernel and the hyperspectral data, we introduce a specialized $1 \times 1$ convolutional layer with $c$ convolution kernels whose weight is parameterized by a Gaussian function, which acts like the spectral sensitivity of visual opsins.

\begin{equation}
\label{eq:gaussian kernel}
\begin{split}
  \psi_c\in\Big\{\Psi\ |\ \Psi = \frac{1}{\sqrt{2 \pi}\sigma}e^{-\frac{(\lambda_i-\lambda_{max,c})^{2}}{2 \sigma^{2}}}, &i\in\{0,1,\cdots,N\}\Big\},\\
  &c\in\{0,1,\cdots,C\}
\end{split}
\end{equation}
As shown in Eq.\ref{eq:gaussian kernel}, $\psi_c$ denotes the $c$-th $1 \times 1$ convolution kernel weight and $\lambda_i$ represents the wavelength corresponding to the $i$-th channel of HSI. $\lambda_{max,c}$ is the maximum sensitive wavelength and $\sigma$ is the standard deviation of the Gaussian function.

\subsection{Evolutionary Conservation Optimization}
\label{sec:optimization}

\subsubsection{MiT Encoder and Segmentation Decoder}
\label{sec:encoder-decoder}
While any encoder architecture could process the opsin layer output \cite{dosovitskiy2021imageworth16x16words, fan2021multiscalevisiontransformers}, we employ Mix Transformers 
\cite{DBLP:journals/corr/abs-2105-15203} for optimal segmentation performance. The opsin output is tokenized into patches via linear projection. As Eq. \ref{eq:encoder} shows, the encoder $E$ generates multi-resolution features $\Phi$, with resolutions decreasing while channel dimensions increase.
\begin{equation}
  E(Input_{i})= \Phi = \{\Phi_{1},\Phi_{2},\Phi_{3},\Phi_{4}\}
  \label{eq:encoder}
\end{equation}

Unlike ViT's single-scale output \cite{dosovitskiy2021imageworth16x16words, chen2023visiontransformeradapterdense}, MiT's hierarchical structure produces CNN-like multi-scale features \cite{DBLP:journals/corr/abs-2105-15203}, providing flexible representations for downstream tasks. This accommodates varying resolution requirements through progressive downsampling ($\frac{H}{2^{i+1}}\times\frac{W}{2^{i+1}}\times C_i$, where $i \in \{1,2,3,4\}$) and channel expansion ($C_{i+1} > C_i$).
And we employ a lightweight all-MLP decoder\cite{DBLP:journals/corr/abs-2105-15203} that utilises all multi-resolution features $\{\Phi_{1},\Phi_{2},\Phi_{3},\Phi_{4}\}$ from the encoder.

\subsubsection{Conservative Regularization}
\label{sec:evo regular}
As illustrated in Fig. \ref{fig:main}.c, direct optimisation of the opsin layer tends to yield unconstrained spectral sensitivities that diverge from biological reality. This discrepancy arises from two fundamental biological constraints: first, opsin sensitivities are typically characterized by Gaussian-like functions centered at the peak absorption wavelength, $\lambda_{max}$ \cite{LOEW19941427, Schott2022, GOVARDOVSKII_FYHRQUIST_REUTER_KUZMIN_DONNER_2000}; second, natural mutations at specific amino acid residues—known as spectral tuning sites (Fig. \ref{fig:main}.b)—primarily induce shifts in $\lambda_{max}$ ranging from 5 to 25 nm rather than altering the overall functional form \cite{YOKOYAMA2000385, spectraltuning1, spectraltuning2}.To bridge this gap, we introduce a conservative regularisation framework to govern the evolution of the opsin layer. Specifically, we parameterize the convolution kernel weights within the opsin layer as a Gaussian function, thereby restricting optimization exclusively to the adjustment of $\lambda_{max}$. To further ensure the optimization trajectory aligns with evolutionary dynamics, we limit the $\lambda_{max}$ shift to a maximum of 0.5 nm per epoch through precise learning rate control, reflecting the incremental nature of natural mutational effects \cite{YOKOYAMA2000385, spectraltuning2}.

Fig.\ref{fig:main} shows evolutionary pressures and selective forces shaping vision arise from factors like food detection, predation avoidance, and social behaviour \cite{carvalho2017genetic}. Furthermore, modern neuroscience \cite{pasupathy2015neural} suggests that semantic segmentation embodies the core challenges of biological vision: bridging low-level perception (textures, edges, and spectral signatures) with high-level semantic reasoning (context, categories, and spatial relationships). Therefore, we employ the segmentation task by utilizing cross-entropy loss on relative datasets. Under our evolutionary conservation optimization, we concurrently train the opsin layer, transformer head, and segmentation decoder on HSI datasets.

\section{Experimental Platform for Evolution}
\label{sec:experiments}
 The number of visual opsins in vertebrate species has undergone two major evolutionary transitions: from four opsins in early vertebrates to two in the mammalian lineage, and from two to three opsins in the primate lineage \cite{evolutionofverte1998}. Interestingly, colour blindness persists in certain primate populations, raising questions about its evolutionary significance \cite{khakinature}. In contrast, deep-sea creatures adapted to low light often possess only rod opsins \cite{deepsearodonly}, with many relying on a single rod opsin \cite{multirod}. Shozo \textit{et al.}~\cite{squirrelfish} proposed the maximum sensitive wavelengths $\lambda_{max}$ of rod opsin in squirrelfishes exhibit a blue-shift trend as living depth increases, as shown in Fig.\ref{fig:blue-shift}.a. Notably, Musilova \textit{et al.}~\cite{multirod} found that certain deep-sea teleost lineages have independently developed multiple rod opsins.

Our analysis begins by reconstructing opsin transitions in mammalian (Sec.~\ref{exp:mammals}) and primate lineages (Sec.~\ref{exp:primates}). We then conduct three key quantitative investigations: 1) the advantages of colour blindness over normal vision (Sec.~\ref{exp:color blind}), 2) the depth-dependent blue-shift phenomenon in squirrelfish opsins and development of multiple rod opsins in deep-sea species (Sec.~\ref{exp:deep sea}), and 3) predictions of colour vision for Martian-adapted creatures (Sec.~\ref{exp:martian envir}). Complete experimental details, including datasets (Sec.~\ref{Exp-Datasets}, Sec.~\ref{HSI-generation}) and noise modeling (Sec.~\ref{sec:realize}), are provided in the appendix.

\begin{figure}[htbp]
  \centering
  \begin{minipage}[t]{0.6\textwidth}
    \centering
    \includegraphics[width=\textwidth]{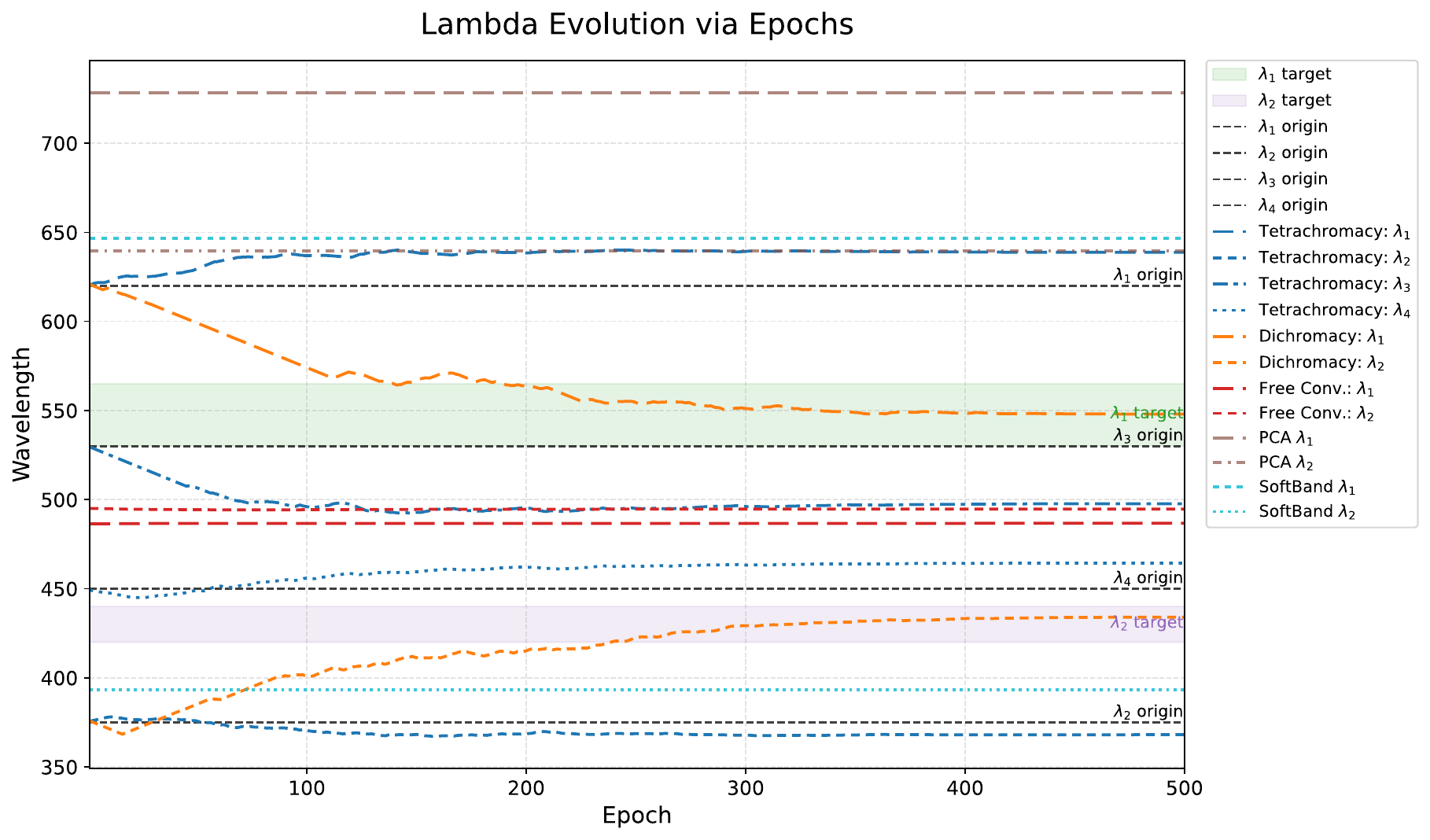}
    \caption{Simulated evolution trajectories of wavelength peak shifts for Tetrachromacy, Dichromacy, and other methods.}
    \label{fig:42}
  \end{minipage}
  \hfill
  \begin{minipage}[t]{0.36\textwidth}
     \centering
    \includegraphics[width=\textwidth]{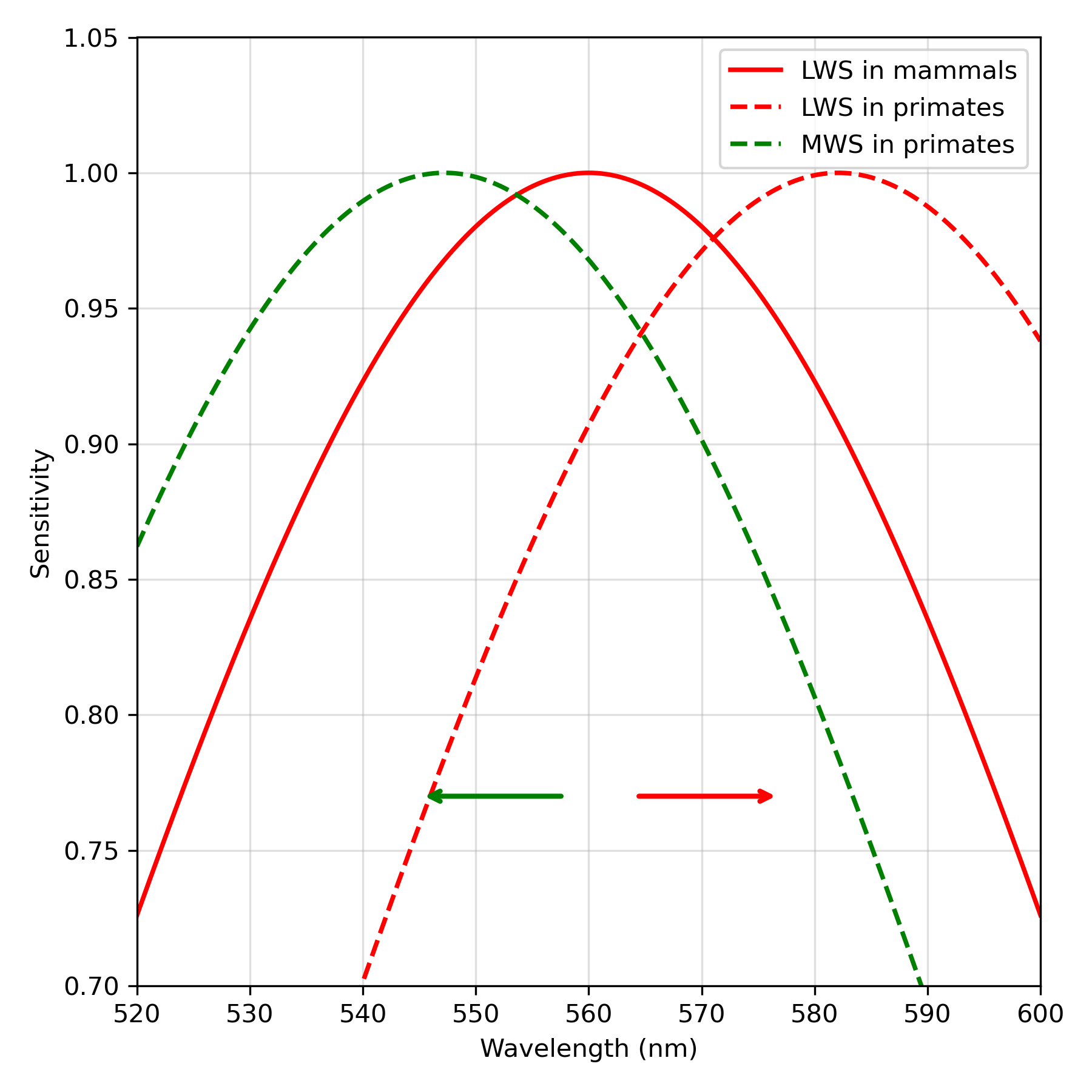}
    \caption{Emulation of the gene duplication process in primates.}
    \label{fig:gene duplication}
  \end{minipage}
\end{figure}

\subsection{Evolutionary Transition from Tetrachromacy to Dichromacy}
\label{exp:mammals}
\vspace{-2pt}

During mammalian evolution, ancestral species experienced a reduction from four to two visual opsin genes \cite{geneloss}, consistent with the nocturnal bottleneck hypothesis \cite{nocturnalbottleneck}. Our experiments model this transition using the four ancestral teleost opsins: LWS (620 nm, $\lambda_1$ origin), SWS1 (375 nm, $\lambda_2$ origin), MWS (530 nm, $\lambda_3$ origin) and SWS2 (450 nm, $\lambda_4$ origin) \cite{evolutionofverte1998}. This framework specifically examines the loss of MWS and SWS2 opsins in the marsupial and eutherian lineages that preceded primates \cite{YOKOYAMA2000385}.
Using the LIB-HSI dataset \cite{habili2023hyperspectral} under dim-light simulation, we modelled dichromatic vision with LWS and SWS1 opsins. As shown in Fig.\ref{fig:42} and Tab. \ref{tab:dichromacy}, our model produced spectral sensitivities consistent with evolutionary predictions and the physiology of extant nocturnal mammals. Through conservation optimization, our in silico evolution achieved a 70nm blue-shift (decrease in wavelength) of LWS and a 60nm red-shift (increase in wavelength) of SWS1 within only 500 epochs, reaching the target wavelength ranges (the $\lambda1$ and $\lambda2$ target intervals in Fig.\ref{fig:42}) of the two opsins in eutherians. We further compared Tetrachromacy versus Dichromacy. Under both fixed (Simulation of the time point of gene loss from Tetrachromacy to Dichromacy) and learnable (Simulation of the respective evolution of Tetrachromacy and Dichromacy) opsin layer conditions, dichromatic vision consistently outperformed tetrachromatic vision in segmentation tasks. Finally, we compared our model with several pure machine learning and statistical methods (see Sec.\ref{ml-methods}). Although those methods achieved performance comparable to our simulated dichromatic evolution, their corresponding peak spectral responses deviated markedly from biological observations, thereby highlighting the biological plausibility of the opsin layer model and the conservation optimization.


\subsection{Evolutionary Transition from Dichromacy to Trichromacy}
\label{exp:primates}

\begin{figure}[htbp]
  \centering

  \begin{minipage}[t]{0.54\textwidth}
    \centering
    \captionof{table}{Tetrachromacy-to-Dichromacy Evolution.}
    \label{tab:dichromacy}
    \vspace{4pt}
    \resizebox{\linewidth}{!}{
      \begin{tabular}{l c c c c c}
        \toprule
        \rowcolor{headerblue}
        \textbf{Methods}
          & \textbf{LWS $\lambda_1$}
          & \textbf{SWS1 $\lambda_2$}
          & \textbf{MWS $\lambda_3$}
          & \textbf{SWS2 $\lambda_4$}
          & \textbf{mIoU} \\
        \rowcolor{headerblue}
          & (nm) & (nm) & (nm) & (nm) & (\%) \\
        \midrule
        \rowcolor{groupgray}
        \multicolumn{6}{l}{\textit{Ours}} \\
        ~~Dichromacy (Fixed)   & 620.00 & 375.00 & --     & --     & 26.95 \\
        ~~Dichromacy (Evolve)  & 548.35 & 433.25 & --     & --     & \textbf{30.49} \\
        ~~Tetrachromacy (Fixed)& 620.00 & 530.00 & 450.00 & 375.00 & 19.93 \\
        ~~Tetrachromacy (Evolve) & 638.85 & 368.05 & 497.60 & 464.30 & 26.33 \\
        \midrule
        \rowcolor{groupgray}
        \multicolumn{6}{l}{\textit{Baselines}} \\
        ~~Free Conv.           & 486.71 & 494.72 & --     & --     & 29.03 \\
        ~~PCA                  & 728.40 & 639.60 & --     & --     & 26.86 \\
        ~~SoftBand             & 646.67 & 393.33 & --     & --     & 27.49 \\
        \bottomrule
      \end{tabular}
    }
  \end{minipage}
  \hfill
  \begin{minipage}[t]{0.44\textwidth}  
    \centering
    \captionof{table}{Dichromacy-to-Trichromacy Evolution via Gene Duplication.}
    \label{tab:trichromacy}
    \vspace{4pt}
    \resizebox{\linewidth}{!}{
      \begin{tabular}{l c c c c c}
        \toprule
        \rowcolor{headerblue}
        \textbf{Methods}
          & \textbf{$\lambda_1$}
          & \textbf{$\lambda_2$}
          & \textbf{$\lambda_3$}
          & \textbf{mIoU}
          & \textbf{$S_R$} \\
        \rowcolor{headerblue}
          & (nm) & (nm) & (nm) & (\%) & \\
        \midrule
        \rowcolor{groupgray}
        \multicolumn{6}{l}{\textit{Ours}} \\
        ~~Dichromacy           & 425.00 & 560.00 & --     & 46.29          & 0.334 \\
        ~~Trichromacy          & 425.00 & 540.00 & 580.00 & 48.45 & \textbf{0.208} \\
        ~~Dichromacy (Evolve)  & 440.65 & 547.34 & 582.10 & 47.99          & 0.239 \\
        \midrule
        \rowcolor{groupgray}
        \multicolumn{6}{l}{\textit{Baselines}} \\
        ~~Free Conv.           & 510.20 & 530.42 & 570.22 & \textbf{49.44}          & 0.224 \\
        ~~PCA                  & 525.10 & 546.25 & 681.43 & 44.79          & 0.312 \\
        ~~SoftBand             & 400.39 & 460.75 & 610.38 & 48.67          & 0.250 \\
        \bottomrule
      \end{tabular}
    }
  \end{minipage}

\end{figure}

The evolution of trichromacy in early primates (~65 MYA) via LWS/MWS opsin gene duplication \cite{nocturnalbottleneck, evolutionofgeneduplication} enabled spectral discrimination critical for fruit/foliage identification \cite{advantageoftri} and camouflage detection \cite{lai2024camoteacherdualrotationconsistencylearning}. This adaptation provided selective advantages for diurnal foraging, establishing the neural basis for red-green colour vision in primates. Accordingly, we performed segmentation tasks using the MinneApple dataset \cite{hani2019minneapple} (Fig. \ref{fig:verte evolution}). Following Lamdouar et al. \cite{lamdouar2023makingbreakingcamouflage}, we quantify fruit recognition using a camouflage score (Sec.\ref{supple:camouflage score}). This metric evaluates how effectively foliage obscures red fruits in segmentation tasks. Lower values indicate better discriminability between leaves and fruits. We simulate trichromatic vision using an opsin layer with $\lambda_{max}$ at 580/540/425 nm, and dichromatic vision by deactivating one channel. Comparative camouflage scores (Tab. \ref{tab:trichromacy}) demonstrate trichromacy's superior fruit recognition, quantitatively validating its evolutionary advantage in foliage foraging.

We modeled the dichromatic-to-trichromatic transition via simulated opsin gene duplication \cite{genedup}, generating 540 nm and 580 nm variants from a single ancestral gene \cite{evolutionofgeneduplication}. This reconstruction used the LIB-HSI dataset \cite{habili2023hyperspectral} to replicate ancestral primate diurnal conditions. Our model initialises opsin layers with 
$\lambda_{1max}$, $\lambda_{2max}$, and $\lambda_{3max}$ at 560, 560, and 425 (fixed) nm. This constrained optimisation simulates both the initial gene duplication and the subsequent spectral divergence. Fig. \ref{fig:gene duplication} demonstrates our model's successful reconstruction of the dichromatic-to-trichromatic transition, providing computational validation of gene duplication's role in colour vision evolution. Moreover, the marked reduction in camouflage scores (Tab. \ref{tab:trichromacy}) further supports the role of gene duplication and its evolutionary process in colour vision evolution. Additionally, we compared our model with other machine learning methods (see Sec. \ref{ml-methods}). The results show that although those methods achieve comparable performance, they cannot align with biological facts. This study specifically models the human trichromatic pathway, not to compare tetrachromatic and trichromatic vision systems. The lack of tetrachromacy re-emergence in our simulations may reflect neutral processes like genetic drift \cite{genedrift}, rather than trichromatic superiority.

\subsection{Retention of Colour Blindness}
\label{exp:color blind}
\begin{wraptable}{r}{0.56\textwidth}
    \vspace{-10pt}
    \centering
    \caption{Fruit recognition ($S_R$) across lighting and colour vision types.
             Lower $S_R$ indicates better camouflage-breaking ability.}
    \label{tab:color_blind}
    \vspace{2pt}
    \resizebox{\linewidth}{!}{
    \begin{tabular}{l ccc cc}
        \toprule        
        \rowcolor{headerblue}
        \textbf{Lighting}
            & \textbf{Normal}
            & \textbf{G.-blind}
            & \textbf{R.-blind}
            & \textbf{G.-weak}
            & \textbf{R.-weak} \\
        \midrule
        Bright      & \textbf{0.208} & 0.316 & 0.333 & 0.254 & 0.237 \\
        Dark (0.10) & 0.309 & 0.391 & 0.365 & 0.293 & \textbf{0.278} \\
        Dark (0.05) & 0.355 & 0.420 & 0.397 & 0.324 & \textbf{0.324} \\
        \bottomrule
    \end{tabular}
    }
    \vspace{-8pt}
\end{wraptable}
The persistence of primates colour blindness remains puzzling. While trichromats have advantages for frugivory in daylight, dichromats may gain compensatory benefits in dark. The leading dim-light advantage hypothesis holds that colour-deficient animals recognise fruit more accurately in near darkness \cite{advantageoftri} and and show a 0.44 log-unit lower light perception threshold after dark adaptation \cite{scotopic1}. However, some researchers found no clear differences from colour-deficient subjects in visual tests \cite{scotopic2}.
Using the MinneApple dataset \cite{hani2019minneapple} and camouflage score, we compared fruit recognition between normal trichromatic vision and simulated colour blindness (via opsin channel modulation). Under dim-light conditions (simulated via noise layer in Sec. \ref{sec:realize}]), colour-weak vision outperformed normal trichromacy in fruit detection (Tab. \ref{exp:color blind}).
Our experimental findings provide substantive support for the leading dim-light advantage hypothesis.

        

\subsection{Deep Sea Creatures}
\label{exp:deep sea}


It is worth mentioning that in this section we simulate underwater data by pre-processing the LIB-HSI dataset \cite{habili2023hyperspectral}\cite{li2024watersplattingfastunderwater3d}. Specifically, we utilise the spectrum of ambient light $E$ in Eq.\ref{eq:spectrum-d} to reduce the intensity distribution of 204 channels based on different wavelengths. Therefore, we simulate the underwater spectrum and intensity of ambient at different depths.
\begin{equation}
    E(d, \lambda)  =  E(0, \lambda) e^{-K_{d}(\lambda) d}
    \label{eq:spectrum-d}
\end{equation}
Here, $d$ represents the depth, $\lambda$ is the wavelength, $E(d, \lambda)$ is the ambient light spectrum at depth $d$, and $E(0, \lambda)$ represents the spectrum at the sea surface. $K_{d}(\lambda)$ is diffuse downwelling attenuation \cite{depthdiffuse}. Additionally, we apply the noise layer (Sec.\ref{sec:realize}) for the dim-light conditions in the deep sea.

\subsubsection{Blue-shift in Rod Opsin of Squirrelfishes}
\label{exp:blue shift}

\vspace{-2pt}

\vspace{-2pt}

\begin{wrapfigure}{r}{0.60\textwidth}
    \vspace{-14pt}
    \centering    \includegraphics[width=\linewidth]{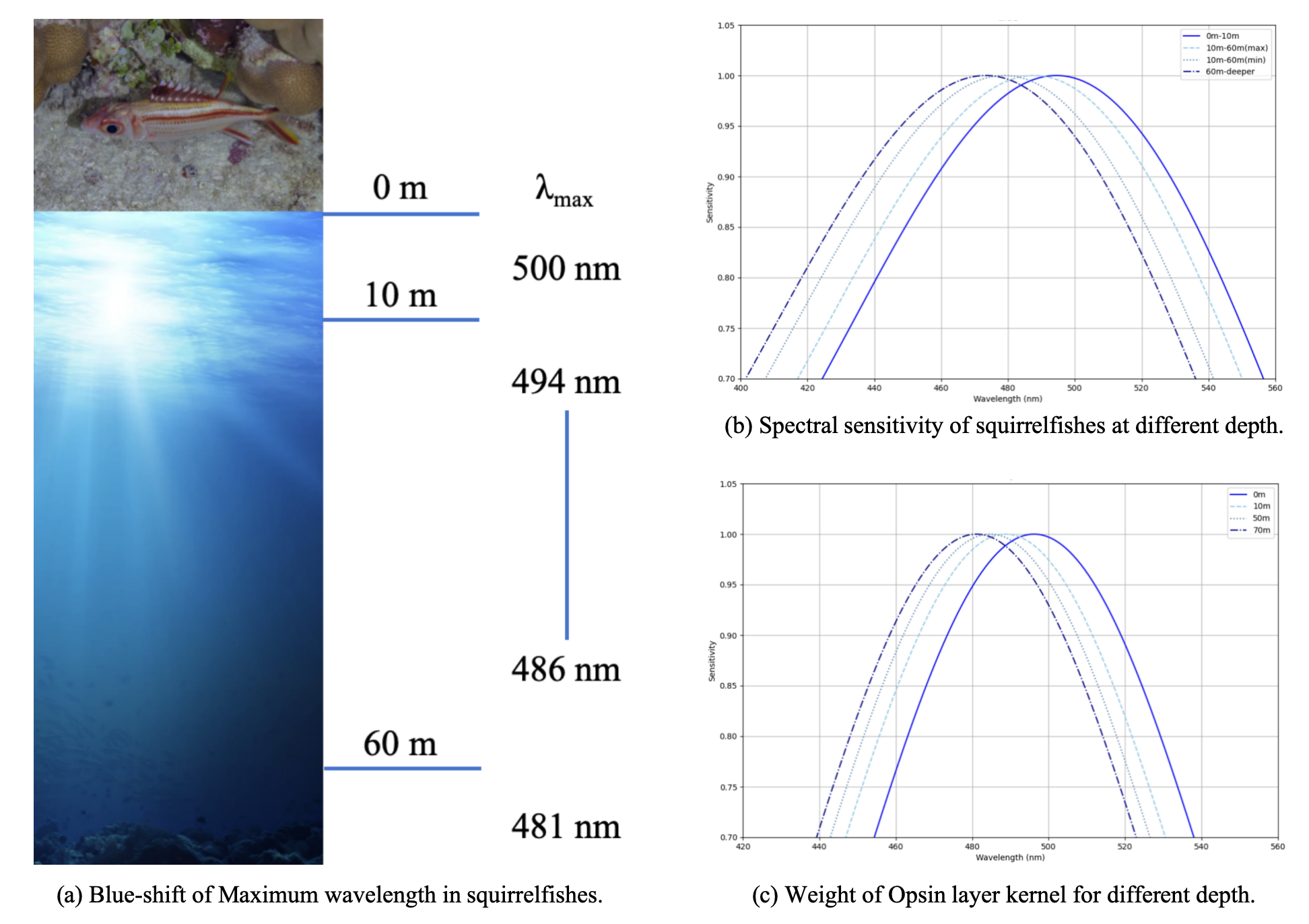}
    \caption{Depth-dependent blue-shift in squirrelfish rod opsins.}
    \label{fig:blue-shift}
    \vspace{-12pt}
\end{wrapfigure}

Fig. \ref{fig:blue-shift}.a–b demonstrates the depth-dependent spectral tuning of squirrelfish rod opsins \cite{squirrelfish}. The nine studied species exhibit a progressive blue-shift in $\lambda_{max}$ (481–502 nm) relative to their ancestral opsin ($\lambda_{max}$ = 493 nm) as depth increases. 

We focus on rod opsin (RH1) in deep-sea adaptation studies for two key reasons. First, deep-sea monochromatic light condition falls below cone detection thresholds, making rod opsin the sole functional photoreceptor \cite{doi:10.1126/science.aav4632}. Second, unlike most teleosts with multiple opsin genes, squirrelfish possess only RH1 \cite{doi:10.1126/science.aav4632, squirrelfish}, providing a genetically simple system
for investigating spectral tuning under extreme selection pressures.
We simulated deep-sea spectral adaptation using the LIB-HSI dataset \cite{habili2023hyperspectral}, processed via Eq.\ref{eq:spectrum-d} for depths (0-70m). A single-opsin layer (initial $\lambda_{max}$ = 493nm, ancestral squirrelfish RH1 \cite{squirrelfish}) performed semantic segmentation, modeling rod opsin functionality across photic zones.
Our quantitative analysis (Tab. \ref{tab:blueshift}) reveals a clear depth-dependent reduction in optimal $\lambda_{max}$ for segmentation mIoU. Fig. \ref{fig:blue-shift}.c reconstructs this rod opsin blue-shift in deep-sea conditions, demonstrating that light diffusion drives the spectral tuning observed in squirrelfishes.
\begin{wraptable}{r}{0.26\textwidth}
    \vspace{-14pt}
    \centering
    \captionof{table}{Blue-shift of $\lambda_{\max}$ vs.\ depth.}
    \label{tab:blueshift}
    \begin{tabular}{l c c}
        \toprule
        \rowcolor{headerblue}
        \textbf{Depth} & \textbf{$\lambda_{\max}$} & \textbf{mIoU} \\
        \rowcolor{headerblue}
                       & (nm)                      & (\%)          \\
        \midrule
        0\,m  & 496.19 & 30.38 \\
        10\,m & 488.73 & 25.88 \\
        50\,m & 484.69 & 16.97 \\
        70\,m & 481.10 & \textbf{14.98} \\
        \bottomrule
    \end{tabular}
    \vspace{-6pt}
\end{wraptable}

\par\medskip
\refstepcounter{subsubsection}
\phantomsection
\addcontentsline{toc}{subsubsection}{\protect\numberline{\thesubsubsection}Multi Rod Opsins: Result of Bioluminescence}
\noindent\rlap{\textbf{\thesubsubsection\quad Multi Rod Opsins: Result of Bioluminescence}}
\par\nobreak\smallskip
\label{exp:multi rods}

While many deep-sea species retain a single rod opsin, certain teleost lineages have independently evolved multiple rod opsins ~\cite{multirod}. This adaptation likely reflects specialized bioluminescent conditions in their ecological niches ~\cite{multirod}.
Our bioluminescent deep-sea model (Sec.\ref{supple:multi rods}) shows that: without bioluminescence, rod opsins exhibit minimal spectral separation; with bioluminescence, opsins diverge by a maximum 5 nm across 5 wavelengths.
However, this separation remains smaller than natural variation, suggesting additional complexity in multi-opsin development. Future studies with more precise anatomical models could clarify these mechanisms.

\subsection{Colour Vision for Hypothetical Martian Life}
\label{exp:martian envir}
We train the opsin layer with different number of kernels to explore the colour vision of hypothetical creatures on Mars, which is close to deuteranopia. We place the experimental results in Sec.\ref{supple:evolution for mars}.
\section{Camera Design}
\label{sec:camera design}
Building on our Visual Evolutionary Analysis and Prediction (Sec. \ref{sec:experiments}), we demonstrate how evolutionary principles can guide task-specific camera design. Using standard Sony camera filters (600/540/460 nm) as initial values and leveraging the mathematical equivalence between color imaging and convolutional operations \cite{nie2018deeply}, we optimized spectral response functions (Fig. \ref{fig:main}) for cancer detection and Martian exploration across three datasets (Sec. \ref{camera:dataset}).

\textbf{Camera for Cancer Detection} Building on Varghese et al.'s \cite{ai-surgery} demonstration of AI's potential to transform surgical care, we adapt our evolutionary optimisation framework to develop specialised cancer-detection cameras using the MHSI Choledoch \cite{cancerdataset} and MODID datasets \cite{modid}. This approach extends AI's medical applications from surgical enhancement to diagnostic imaging optimisation.
Our optimised spectral filters (Tab. \ref{tab:cancer}) enhance performance across distinct cancer detection tasks, enabling specialized medical imaging for abnormal tissue identification. Sec. \ref{supple:camera for cancer} presents scalable designs with varying filter counts for different clinical applications.


\textbf{Camera for Martian Environment} Using the Mars-Seg dataset \cite{marsdataset}, we optimized spectral filter responses via opsin-layer training. In segmentation tasks, our optimized filter designs (Tab. \ref{tab:mars}) exhibit superior performance, with extended multi-filter configurations detailed in Sec. \ref{supple:camera for mars}. These findings suggest that such optimized vision systems could substantially enhance the perceptual capabilities of Mars rovers and exploration vehicles.

In Tab. \ref{tab:cancer} and \ref{tab:mars}, we first benchmark our approach against established baselines for Hyperspectral Image processing. A longitudinal review of baseline methods from 2015 to 2025 reveals that annual performance gains have averaged only 1\%; in contrast, our approach first achieves state-of-the-art performance based on the MiT-Transformer and further yields an improvement of over 2\% mIoU after learning through the opsin layer. We observe that while conventional methods—which focus primarily on iterative improvements to high-level perceptual algorithms—show diminishing returns and slow performance gains, our perspective of low-level camera design and simple hardware modifications yields a substantial performance leap. This aligns with previous findings \cite{flexisp, differentiable} and offers a new perspective on task-oriented camera design.
Furthermore, we evaluate our framework across three configurations: General (non-optimized filters), Free (unconstrained filter optimization), and our proposed Design (progressive, conservative optimization via the Opsin Layer). The results demonstrate that the Design method not only achieves peak performance but also maintains spectral parameters ($\lambda$) that remain close to original hardware specifications. This "conservative optimization" is of significant practical value, as it ensures that the resulting designs are biologically grounded and feasible for 
iterative manufacturing and real-world camera deployment.
\begin{table*}[htbp]
\centering
\caption{Camera design for bile duct and oral cancer detection (MHSI and MODID datasets).}
\label{tab:cancer}
\resizebox{0.86\textwidth}{!}{
\begin{tabular}{ll c ccc c}
    \toprule
    \rowcolor{headerblue}
    \textbf{Dataset} & \textbf{Methods} & \textbf{Param (M)}
        & \textbf{$\lambda_{1}$ (nm)} & \textbf{$\lambda_{2}$ (nm)} & \textbf{$\lambda_{3}$ (nm)}
        & \textbf{mIoU (\%)} \\
    \midrule
    \rowcolor{groupgray}
    \multicolumn{7}{l}{\textit{Baselines (MHSI / MODID)}} \\
    ~~MHSI/MODID & UNet (2015)\cite{unet}        & 31.07  & -- & -- & -- & 51.04 / 54.02 \\
                 & SwinUNet (2022)\cite{swinunet} & 41.48  & -- & -- & -- & 47.84 / 51.12 \\
                 & ResUNet (2020)\cite{resunet}   & 13.11  & -- & -- & -- & 49.52 / 52.76 \\
                 & Segformer (2021)\cite{segformer}& 64.47 & -- & -- & -- & 50.27 / 52.98 \\
                 & EMCAD (2024)\cite{emcad}       & 84.53  & -- & -- & -- & 50.07 / 52.67 \\
                 & nnUNet (2021)\cite{nnunet}     & 19.68  & -- & -- & -- & 49.99 / 52.57 \\
                 & MADGNet (2024)\cite{madgnet}   & 31.39  & -- & -- & -- & 51.68 / 54.45 \\
                 & H-VMUNet (2025)\cite{h-vmunet} & 117.49 & -- & -- & -- & 51.30 / 52.71 \\
                 & UDMamba (2024)\cite{udmamba}   & 19.21  & -- & -- & -- & 50.89 / 53.81 \\
                 & FDNet (2024)\cite{fdnet}       & 96.26  & -- & -- & -- & 53.29 / 56.21 \\
                 & Spectr (2024)\cite{spectr}     & 3.94   & -- & -- & -- & 52.19 / 55.03 \\
                 & UFPF (2025)\cite{ufpf}         & 201.35 & -- & -- & -- & 54.99 / 56.98 \\
    \midrule
    \rowcolor{groupgray}
    \multicolumn{7}{l}{\textit{Ours}} \\
    ~~MHSI  & General & 15.10 & 600.00 & 540.00 & 460.00 & 54.35 \\
            & Free    & 15.10 & 689.25 & 716.45 & 691.44 & 54.73 \\
            & Design  & 15.10 & 595.82 & 716.45 & 476.60 & \textbf{56.36} \\
    \midrule
    ~~MODID & General & 15.10 & 600.00 & 540.00 & 460.00 & 56.42 \\
            & Free    & 15.10 & 542.58 & 559.25 & 558.86 & 57.69 \\
            & Design  & 15.10 & 608.73 & 552.93 & 449.54 & \textbf{58.67} \\
    \bottomrule
\end{tabular}}
\end{table*}

\begin{center}
    \centering
    
\captionof{table}{Camera design for Martian environment (Mars-Seg).}
\label{tab:mars}
\resizebox{0.86\textwidth}{!}{
\begin{tabular}{l c c ccc c}
    \toprule
    \rowcolor{headerblue}
    \textbf{Methods} & \textbf{Input} & \textbf{Param (M)}
        & \textbf{$\lambda_{1}$ (nm)} & \textbf{$\lambda_{2}$ (nm)} & \textbf{$\lambda_{3}$ (nm)}
        & \textbf{mIoU (\%)} \\
    \midrule
    \rowcolor{groupgray}
    \multicolumn{7}{l}{\textit{RGB Baselines}} \\
    ~~UNet (2015)\cite{unet}         & RGB & 31.07 & -- & -- & -- & 50.65 \\
    ~~SwinUNet (2022)\cite{swinunet} & RGB & 41.48 & -- & -- & -- & 64.74 \\
    ~~Segformer (2021)\cite{segformer}& RGB & 64.47 & -- & -- & -- & 62.36 \\
    \midrule
    \rowcolor{groupgray}
    \multicolumn{7}{l}{\textit{HSI Baselines}} \\
    ~~UNet (2015)\cite{unet}         & HSI & 31.07  & -- & -- & -- & 35.03 \\
    ~~SwinUNet (2022)\cite{swinunet} & HSI & 41.48  & -- & -- & -- & 36.44 \\
    ~~ResUNet (2020)\cite{resunet}   & HSI & 13.11  & -- & -- & -- & 35.51 \\
    ~~Segformer (2021)\cite{segformer}& HSI & 64.47 & -- & -- & -- & 36.93 \\
    ~~UDMamba (2024)\cite{udmamba}   & HSI & 19.21  & -- & -- & -- & 38.55 \\
    ~~Spectr (2024)\cite{spectr}     & HSI & 3.94   & -- & -- & -- & 39.17 \\
    ~~UFPF (2025)\cite{ufpf}         & HSI & 201.35 & -- & -- & -- & 40.14 \\
    \midrule
    \rowcolor{groupgray}
    \multicolumn{7}{l}{\textit{Ours}} \\
    ~~General & HSI & 15.10 & 600.00 & 540.00 & 460.00 & 39.22 \\
    ~~Free    & HSI & 15.10 & 724.03 & 712.25 & 423.43 & 40.18 \\
    ~~Design  & HSI & 15.10 & 623.35 & 580.41 & 489.13 & \textbf{41.23} \\
    \bottomrule
\end{tabular}}
\end{center}

\clearpage
\section{Summary}
\label{sec:conclusion}
In this paper, we present a paleoinspired vision framework that models retinal visual transduction and simulates evolutionary transitions of visual opsins. Our approach enables the quantitative analysis of opsin evolution with high computational efficiency, addressing a core challenge in evolutionary biology and neuroscience. While methods like forward genetics experiments often suffer from resource-intensive stochastic searches for difficult-to-quantify traits \cite{Hawkins2021}, our framework provides algorithmic tools to streamline biological research. By implementing processes like evolutionary pressure analysis, we can quantify the direction and magnitude of trait changes in successive generations and simulate the specific environmental conditions of extinct species within an optimization-based paradigm.
Beyond theoretical insights, our framework establishes practical paradigms for high-performance camera design. The evolutionary conservatism embedded in our optimization supports manufacturing-friendly filter designs \cite{manufacturing}, while our minimalist approach, maximizing task-specific adaptability, aligns with emerging trends in computational imaging \cite{Minimalist_ECCV24}. 

\begin{ack}
This work was partly supported by JSPS KAKENHI Grant Number JP26K21296 (to Z.~C.), 23K23924 and 24K21963 (to D.K.), Cooperative Research Project of the Research Institute of Electrical Communication, Tohoku University (to R.~F.), and the National Key R\&D Program of China Grant Number 2025YFF0811700 (to J.~L.).
\end{ack}

\bibliographystyle{abbrv}
\bibliography{main}


\appendix

\setcounter{tocdepth}{4}
\setcounter{secnumdepth}{4}
\clearpage

\section{Technical Appendices and Supplementary Material}
\subsection{Implementation Details}
\label{suppl:details}
\subsubsection{Layers after Opsin Layer}

\begin{figure}[htbp]
 \centering
 \includegraphics[width=1.0\linewidth]{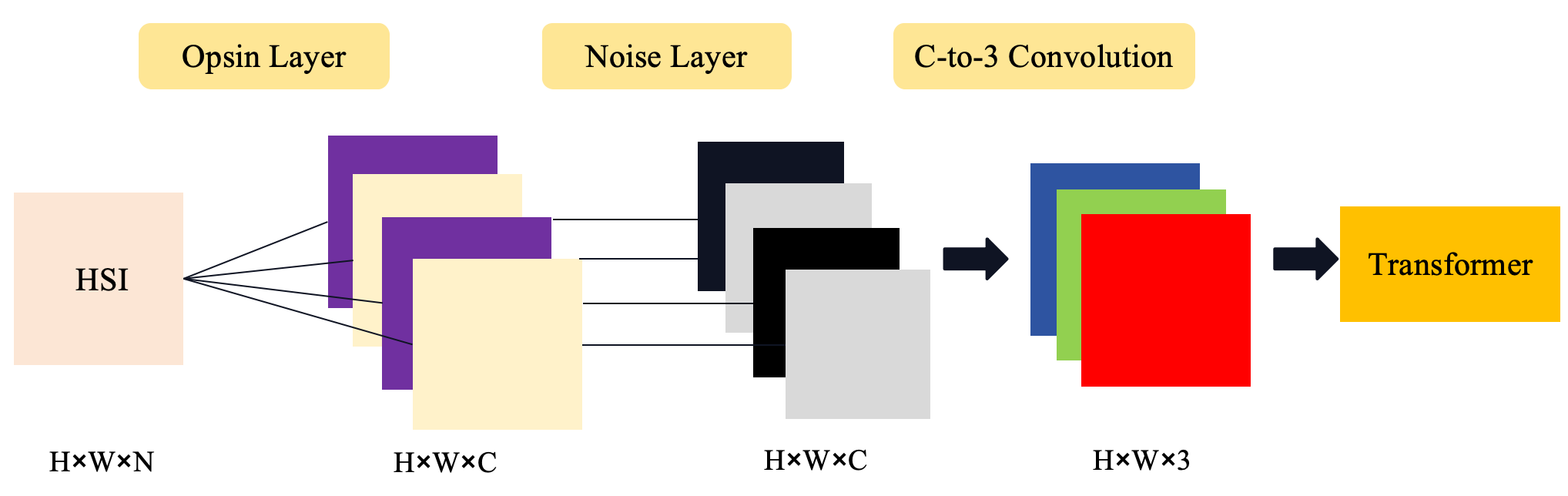}
 \vspace{-3mm}
 \caption{Architecture of the network.}
 \vspace{-3mm}
 \label{fig:archi of SRE}
\end{figure}

\textbf{Noise Layer}
The noise layer is applied after each channel of the opsin layer's output to incorporate the noise present in opsins. Let the output of the opsin layer for channel $c$ be $I_c$; the output of the noise layer is expressed as:
\begin{equation}
    I_c^{noise} = I_c + Gaussian(\mu = 0,\ \sigma^2 = \tau\times I_c)
\end{equation}
where $Gaussian(\mu,\sigma^2)$ represents Gaussian noise with a mean $\mu$ and variance $\sigma^2$, and $\tau$ is a scaling factor that determines the noise intensity as illustrated in Sec.\ref{sec:realize}.

\textbf{C-to-3 Convolution Layer.}
The MiT encoder used in our work requires input data of size $H\times W \times 3$, corresponding to a 3-channel image. However, the output of our opsin layer is of size $H\times W\times C$, where $C$ represents the number of convolutional kernels in the opsin layer, corresponding to the simulated number of opsins. To ensure compatibility with the MiT encoder, we introduce a $1\times 1$ convolutional layer to transform the feature map from $H\times W\times C$ to $H\times W\times 3$. As shown in Fig.\ref{fig:archi of SRE}, this $1\times 1$ convolution layer is applied after the opsin layer and the noise layer.

\textbf{Training Stage.}
In the training stage, we only train the opsin layer and C-to-3 convolution layer (if exist) simultaneously with the encoder and decoder, while fixing the noise layer (if exist).
\begin{figure}[htbp]
 \centering
 \includegraphics[width=1.0\linewidth]{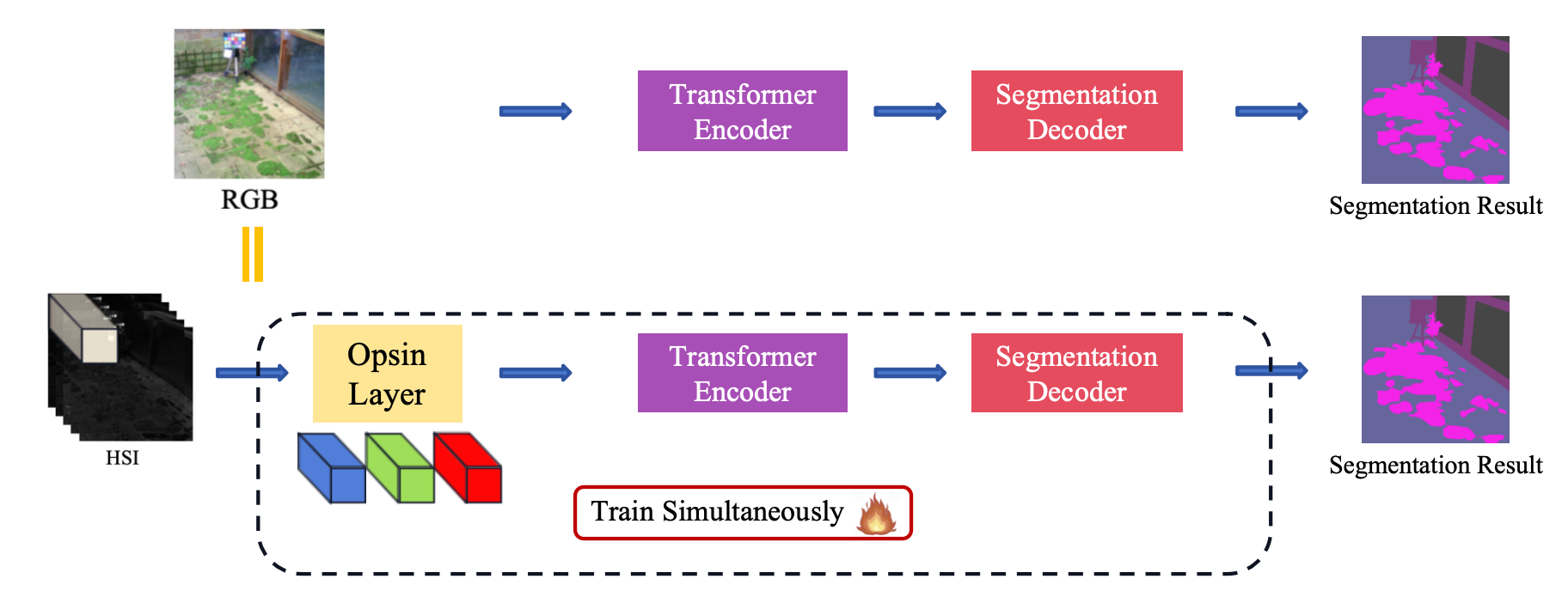}
 \vspace{-3mm}
 \caption{Training stage with Opsin Layer, Transformer Encoder and Segmentation Decoder.}
 \vspace{-3mm}
 \label{fig:training stage}
\end{figure}

\subsubsection{Camouflage Score}
\label{supple:camouflage score}
Specifically, we utilize the reconstruction fidelity score proposed by Hala Lamdouar \textit{et al.} \cite{lamdouar2023makingbreakingcamouflage} as the metric to evaluation the ability of the colour vision to recognize camouflage. For the LMS image $F$ output from the opsin layer and segmentation mask $m$, we first get the foreground region and the background region by erosion and dilation operations on the mask, 
\begin{equation}
    F_{fg}, F_{bg} = \text{erode}(m)\odot F, (1-\text{dilate}(m))\odot F
\end{equation}
We calculate the reconstruction fidelity score, which is, briefly, the proportion of pixels in the foreground that can be successfully reconstructed from the background \cite{lamdouar2023makingbreakingcamouflage},
\begin{equation}
    S_R = \frac{1}{N_{fg}}\sum_{(i,j)\in F_{fg}} R_f(i,j)
\end{equation}
\begin{equation}
    R_f(i,j) = \Big[||F_{fg}-\Phi(F_{fg})||_2 < t ||F_{fg}||_2\Big]
\end{equation}
where $N_{fg}$ is the number of pixels in $F_{fg}$, $\Phi$ is the reconstruction operation and $t$ is a threshold parameter. A lower $S_R$ value indicates a superior capability of the model to break camouflage, suggesting that the simulated animal vision is better adapted to survival pressures. In Sec.\ref{exp:color blind}, we adjust the value of $t$ based on the lighting conditions, as $||F_{fg}||_2$ is small in dim light, making the reconstruction criterion for the foreground region more stringent. Specifically, we set $t=0.2$ for bright conditions, $t=1.2$ for dark (0.1), and $t=1.6$ for darker conditions (0.05). 

\subsubsection{Data Preparation}
\label{data-prep}
\paragraph{Datasets Used in Experimental Platform for Evolution}
\label{Exp-Datasets}
\begin{itemize}
    \item \textbf{LIB-HSI} ~\cite{habili2023hyperspectral} comprises hyperspectral images and corresponding RGB images of building façades, incorporating associated vegetation, bare soil, and ground surfaces. This composition provides critical spectral-spatial features that effectively simulate key environmental parameters required for our experiments. 
    \item \textbf{MinneApple} \cite{hani2019minneapple} It comprises various apple tree species, closely matching the fruit-picking conditions in our experiment. We selected red fruits and extracted image regions containing only leaves and fruits for analysis. 
    
    \item \label{mars-Datasets} \textbf{Mars-Seg} \cite{marsdataset} contains downsampled (560 × 500 px, bilinear interpolation) Mars surface images from NASA's Curiosity rover, representing the most comprehensive and authoritative collection of diverse Martian terrains for scientific research. 
\end{itemize}

\paragraph{Datasets Used in Camera Design}
\label{camera:dataset}
\begin{itemize}
    \item \textbf{Mars-Seg} See Sec. \ref{mars-Datasets}.
    \item \textbf{MHSI Choledoch} \cite{cancerdataset} is a comprehensive collection of microscopy hyperspectral and colour images focused on cholangiocarcinoma, a type of bile duct cancer. 
    \item \textbf{MODID} \cite{modid} is a in-vivo multispectral oral cancer dataset comprising 243 images with 16 spectral bands from 460–600 nm.
\end{itemize}

\paragraph{HSI Generation}
\label{HSI-generation}
Due to the scarcity of high-quality annotated HSI datasets, we generate synthetic HSI data from RGB images using the state-of-the-art method described in \cite{9878871}. This approach is applied to MinneApple and Mars-Seg datasets, enabling us to model photoreceptor spectral sensitivities across diverse real-world scenarios.

\paragraph{Noise of Eyes in Dim-Light Environment}
\label{sec:realize}

We simulate dim-light environments in Sec.\ref{exp:mammals}, Sec.\ref{exp:color blind} and Sec.\ref{exp:deep sea} by adding a noise layer after opsin layer. We simplify the noise in eye as the Poisson fluctuations of dark light \cite{Noise1}\cite{noisesource}. The variance of noise $\sigma^2_{N}$ is shown in Eq.\ref{eq:noise}, where $I$ is the intensity of dark light and $\tau$ is the noise factor. We referred the value of the signal-to-noise ratio (SNR) under dim light for single cone cell in human and rod cell in macaque monkeys measured by H.Richard Blackwell \cite{Noise2} and D.M.Schneeweis \textit{et al.} \cite{snrofrod} correspondingly, hence set the noise factor $\tau=0.1$ for cone cells in Sec.\ref{exp:mammals}, Sec.\ref{exp:color blind} and $\tau=0.02$ for rod cells in Sec.\ref{exp:deep sea}.
\begin{equation}
    \sigma^2_{N}  =  \tau \times I
    \label{eq:noise}
\end{equation}

\subsubsection{Ablation Method Details}
\label{ml-methods}
Let $\mathbf{X}\in\mathbb{R}^{B\times H\times W}$ denote a hyperspectral image with
$B$ spectral bands. The center wavelength of the $b$-th band is denoted as
$\lambda_b$. All compared modules convert
$\mathbf{X}$ into a three-channel representation
$\mathbf{Y}\in\mathbb{R}^{3\times H\times W}$ before it is fed into the downstream
segmentation network (MiT Encoder).
\paragraph{Fixed  Opsin Layer.}
The proposed opsin layer represents each photoreceptor-like channel by a Gaussian spectral
response centered at a biologically motivated peak wavelength $\lambda_c^{\max}$.
For the three-kernel opsin layer, the three peak wavelengths are fixed. The response of the $c$-th channel to the $b$-th band is
\begin{equation}
g_{c,b}
=
\frac{1}{\sigma\sqrt{2\pi}}
\exp\left(
-\frac{(\lambda_b-\lambda_c^{\max})^2}{2\sigma^2}
\right),
\end{equation}
where $\sigma$ nm is fixed. The three-channel output is obtained by a spectral
$1\times1$ convolution:
\begin{equation}
Y_{c,h,w}
=
\sum_{b=1}^{B} g_{c,b} X_{b,h,w}.
\end{equation}
This design imposes an explicit smoothness and unimodality prior on the spectral
responses, mimicking opsin-like sensitivity curves.

\paragraph{Free Spectral Convolution.}
We further consider a fully data-driven spectral mixing baseline implemented as an
unconstrained $1\times1$ convolution from the B bands to three channels:
\begin{equation}
Y_{c,h,w}
=
\sum_{b=1}^{B} W_{c,b}X_{b,h,w},
\end{equation}
where $W\in\mathbb{R}^{3\times B}$ is learned directly from the segmentation loss.
This baseline contains only $3*B$ trainable spectral parameters and
does not assume Gaussianity, smoothness, or biological interpretability.
Although a free convolution does not explicitly define $\lambda_{\max}$, we extract
an equivalent wavelength from the learned spectral response. Since raw convolution
weights may be negative, we first map them to non-negative response values:
\begin{equation}
r_{c,b}
=
\operatorname{softplus}(W_{c,b})
=
\log(1+\exp(W_{c,b})).
\end{equation}
The hard peak wavelength is then defined as
\begin{equation}
\lambda^{\mathrm{peak}}_c
=
\lambda_{\arg\max_b r_{c,b}}.
\end{equation}

\paragraph{PCA Projection.}
We include PCA as a classical non-learnable band-compression baseline. PCA is fitted
only on the training split to avoid information leakage. Specifically, spectral
pixels are sampled from the training hyperspectral cubes and reshaped into a matrix
\begin{equation}
\mathbf{S}
=
[\mathbf{x}_1,\mathbf{x}_2,\ldots,\mathbf{x}_N]^\top
\in\mathbb{R}^{N\times B},
\end{equation}
where each $\mathbf{x}_n\in\mathbb{R}^{B}$ is a spectral vector at one spatial
location. PCA estimates the leading eigenvectors of the empirical covariance matrix
\begin{equation}
\mathbf{C}
=
\frac{1}{N-1}
(\mathbf{S}-\bar{\mathbf{S}})^\top
(\mathbf{S}-\bar{\mathbf{S}}),
\end{equation}
and keeps the first three principal components:
\begin{equation}
\mathbf{P}
=
[\mathbf{u}_1,\mathbf{u}_2,\mathbf{u}_3]^\top
\in\mathbb{R}^{3\times B}.
\end{equation}
Each hyperspectral pixel is then projected to a three-dimensional representation:
\begin{equation}
\mathbf{y}_{h,w}
=
\mathbf{P}\mathbf{x}_{h,w}.
\end{equation}
In our implementation, the PCA matrix is estimated from the training set and reused
unchanged for the validation/test set, ensuring that all splits share the same
projection space. The projected channels are min--max normalized independently.
This baseline has no trainable parameters in the segmentation model and evaluates
whether a variance-preserving linear projection is sufficient for downstream
semantic segmentation.
For interpretability, the spectral characteristics of the $i$-th principal component
can be summarized by the magnitude of its loading vector. Its dominant wavelength is
\begin{equation}
\lambda^{\mathrm{peak}}_i
=
\lambda_{\arg\max_b |P_{i,b}|},
\end{equation}
\paragraph{Gumbel-Softmax Differentiable Band Selection.}
Hard band selection is a natural baseline for hyperspectral compression because it
preserves real sensor bands rather than synthesizing channels by dense spectral
mixing. However, directly selecting bands with an $\arg\max$ operator is
non-differentiable and cannot be optimized end-to-end using gradient descent. We
therefore use a Gumbel-Softmax relaxation, which provides a differentiable
approximation to categorical band selection during training while converging toward
discrete one-hot choices as the temperature decreases.
For each output channel $c$, we maintain learnable logits
$\boldsymbol{\alpha}_c\in\mathbb{R}^{B}$. During training, a relaxed band-selection
vector is sampled as
\begin{equation}
q_{c,b}
=
\frac{
\exp((\alpha_{c,b}+\gamma_{c,b})/\tau)
}{
\sum_{b'=1}^{B}
\exp((\alpha_{c,b'}+\gamma_{c,b'})/\tau)
},
\end{equation}
where $\tau$ is the temperature and $\gamma_{c,b}$ is Gumbel noise sampled by
\begin{equation}
\gamma_{c,b}
=
-\log(-\log u_{c,b}),\qquad
u_{c,b}\sim\mathcal{U}(0,1).
\end{equation}
The selected three-channel representation is computed by
\begin{equation}
Y_{c,h,w}
=
\sum_{b=1}^{B} q_{c,b}X_{b,h,w}.
\end{equation}
The temperature is annealed after each epoch:
\begin{equation}
\tau_t
=
\max(\tau_{\min}, \tau_0\exp(-\eta t)),
\end{equation}
where $\tau_0=1.0$, $\tau_{\min}=0.1$, and $\eta=0.003$ in our implementation.
At inference time, the relaxed distribution is replaced by a hard one-hot selection:
\begin{equation}
b_c^\star
=
\arg\max_b \alpha_{c,b},
\qquad
q_{c,b}
=
\mathbb{I}[b=b_c^\star].
\end{equation}
The corresponding selected wavelength is therefore
\begin{equation}
\lambda_c^{\max}
=
\lambda_{b_c^\star}.
\end{equation}
This baseline is reasonable because it bridges classical discrete band selection and end-to-end neural optimization: during training it remains differentiable, while during inference it yields an interpretable subset of actual MODID wavelengths.

\subsubsection{Training Details}
We have eight RTX 3090 GPUs to run the experiments. We train the segmentation task on various datasets by minimizing the cross-entropy loss and using the Adam optimizer \cite{kingma2017adammethodstochasticoptimization} to update the parameters of the opsin layer, encoder, and decoder.

In training to reconstruct the evolutionary transition to dichromatic vision in mammals (Sec.\ref{exp:mammals}), we use the Adam optimizer to train the opsin layer with a base learning rate of $2\times10^{-2}$ and the encoder and decoder with a base learning rate of $5\times10^{-4}$. Additionally, a cosine learning schedule is applied to mitigate over-fitting.

In training to simulate the gene duplication process in primates leading to the acquisition of trichromatic vision (Sec.\ref{exp:primates}), we use the Adam optimizer to train the opsin layer with a base learning rate of $5\times10^{-2}$ and the encoder and decoder with a base learning rate of $5\times10^{-4}$. Additionally, a cosine learning schedule is applied to mitigate over-fitting.

In training to compare the performance of colour blindness and normal trichromatic vision (Sec.\ref{exp:color blind}), we use the Adam optimizer to train the opsin layer, the encoder and decoder all with a base learning rate of $5\times10^{-4}$. Additionally, a cosine learning schedule is applied to mitigate over-fitting.

In training to reconstruct the blue-shift phenomenon of rod opsins in squirrelfishes (Sec.\ref{exp:blue shift}), we use the Adam optimizer to train the opsin layer with a base learning rate of $5\times10^{-3}$ and the encoder and decoder with a base learning rate of $5\times10^{-4}$. Additionally, a cosine learning schedule is applied to mitigate over-fitting.

In training to quantitatively analyse the evolutionary development of multiple rod opsins in certain deep-sea species (Sec.\ref{exp:multi rods}), we use the Adam optimizer to train the opsin layer with a base learning rate of $1\times10^{-2}$ and the encoder and decoder with a base learning rate of $5\times10^{-4}$. Additionally, a cosine learning schedule is applied to mitigate over-fitting.

To quantitatively analyse the potential colour vision of hypothetical creatures adapted to Martian environments (Sec.\ref{exp:martian envir}), we use the Adam optimizer to train the opsin layer with a base learning rate of $5\times10^{-2}$ and the encoder and decoder with a base learning rate of $5\times10^{-4}$. Additionally, a cosine learning schedule is applied to mitigate over-fitting.

For the camera design section (Sec.\ref{sec:camera design}), we use the Adam optimizer to train the opsin layer with a base learning rate of $3\times10^{-2}$ and the encoder and decoder with a base learning rate of $6\times10^{-5}$. Furthermore, we incorporate an adaptive training strategy, inspired by \cite{qnet}, to dynamically optimize the wavelengths during model training. This approach enables the network to learn task-specific spectral sensitivities, enhancing feature discriminability for oral cancer detection.

The reported results represent mean performance over multiple independent random seeds. The default training duration is 500 epochs; the multi-rod trajectories below are reported through 200 epochs, and the hypothetical Martian-vision trajectories through 1000 epochs. These reported durations should not be interpreted as a common stopping rule, and mean scores alone do not quantify uncertainty.

\subsection{Supplementary Results for Evolution}
\label{supple:evolution}
\subsubsection{Multi Rod Opsins: Result of Bioluminescence}
\begin{figure}[htbp]
 \centering
 \includegraphics[width=0.6\linewidth]{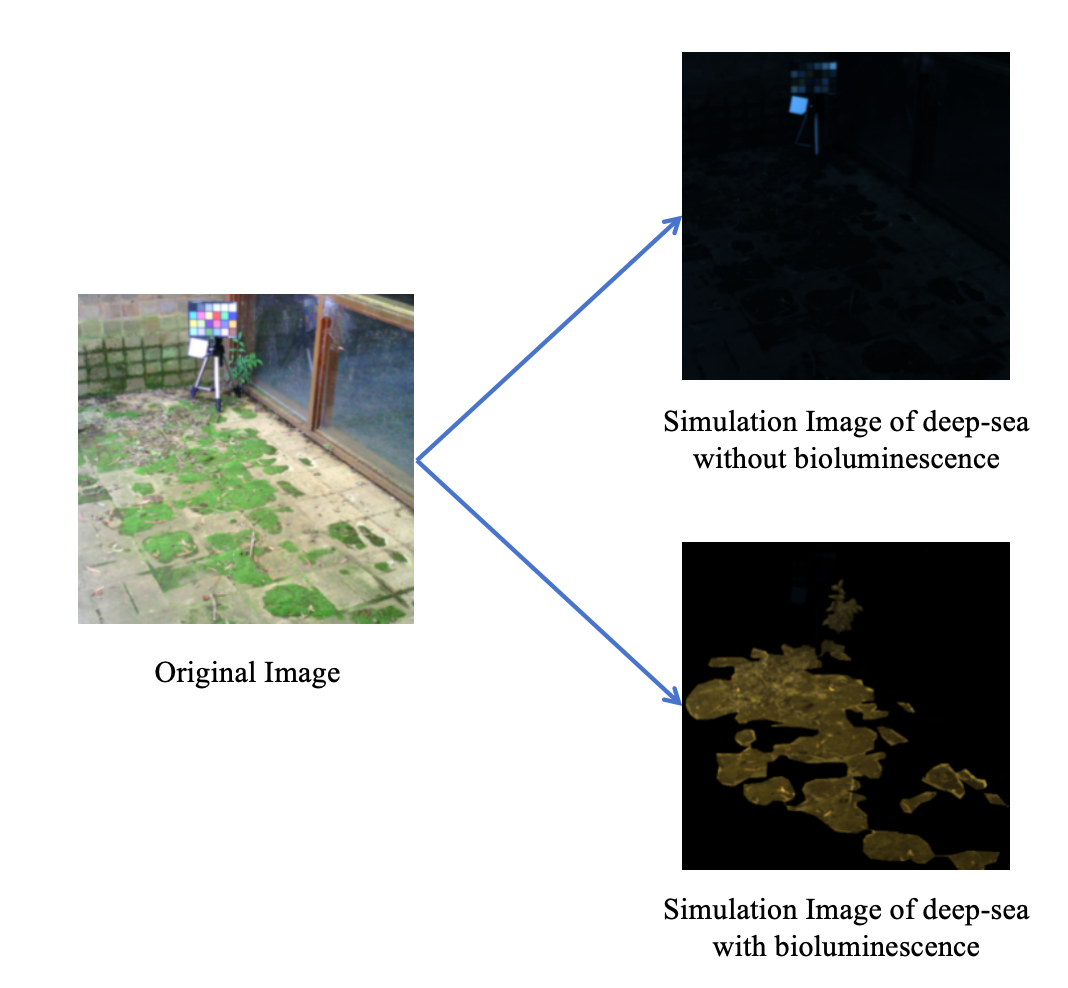}
 \caption{Environment simulation with and without bioluminescence.}
 \label{fig:biolumi simu}
\end{figure}

As shown in Figure~\ref{fig:biolumi simu}, we simulate a deep-sea environment with bioluminescence by retaining the labelled region without underwater attenuation. At epoch 200, the five peak wavelengths span 0.33~nm without bioluminescence (Table~\ref{tab:5 rods without}) and 3.98~nm with bioluminescence (Table~\ref{tab:5 rods with}). The separation is qualitatively consistent with the multi-rod hypothesis~\cite{multirod}, but its magnitude remains smaller than in the biological observations.

This provides quantitative evidence supporting the role of bioluminescence in the development of multiple rod opsins. However, the separation between rod opsins remains minimal compared to biological experimental data \cite{multirod}, aligning with hypotheses suggesting the involvement of more complex mechanisms in the evolution of multiple rod opsins \cite{multirod}.
\label{supple:multi rods}

\begin{table*}[htbp]
    \centering
    \begin{minipage}{0.4\textwidth}
        \centering
        \caption{1-kernel opsin layer without bioluminescence.}
        \label{tab:single rod without}
        \begin{tabular}{c|c|c}
            \hline
            Epoch & $\lambda_{max}$  & mIoU \%   \\ \hline
            0   & 481.00  & - \\ \hline
            50   & 471.86  & 6.69 \\ \hline
            100   & 466.09  & 7.88 \\ \hline
            150   & 461.62  & 10.17 \\ \hline
            200   & 457.23  & 11.14 \\ \hline
        \end{tabular}
    \end{minipage}%
    \hspace{0.3cm} 
    \begin{minipage}{0.4\textwidth}
        \centering
        \caption{1-kernel opsin layer with bioluminescence.}
        \label{tab:single rod with}
        \begin{tabular}{c|c|c}
            \hline
            Epoch & $\lambda_{max}$  & mIoU \%   \\ \hline
            0   & 481.00  & - \\ \hline
            50   & 482.66  & 8.26 \\ \hline
            100   & 482.13  & 10.09 \\ \hline
            150   & 481.41  & 12.01 \\ \hline
            200   & 480.83  & 15.17 \\ \hline
        \end{tabular}
    \end{minipage}%
\end{table*}

\begin{table*}[htbp]
    \centering
    \caption{5-kernel opsin layer without bioluminescence.}
    \label{tab:5 rods without}
    \begin{tabular}{c|ccccc|c}
        \hline
        Epoch & $\lambda_{1max}$ & $\lambda_{2max}$ & $\lambda_{3max}$ & $\lambda_{4max}$ & $\lambda_{5max}$ & mIoU \%   \\ \hline
        0   & 481.00 & 481.00 & 481.00 & 481.00 & 481.00 & - \\ \hline
        50   & 475.41 & 475.41 & 475.26 & 475.56 & 475.29 & 8.44 \\ \hline
        100   & 472.89 & 472.89 & 472.74 & 473.04 & 472.76 & 12.16 \\ \hline
        150   & 470.40 & 470.38 & 470.24 & 470.56 & 470.26 & 13.02 \\ \hline
        200   & 467.76 & 467.73 & 467.59 & 467.92 & 467.62 & 13.77 \\ \hline
    \end{tabular}
\end{table*}
\vspace{-5mm}
\begin{table*}[htbp]
    \centering
    \caption{5-kernel opsin layer with bioluminescence.}
    \label{tab:5 rods with}
    \begin{tabular}{c|ccccc|c}
        \hline
        Epoch & $\lambda_{1max}$ & $\lambda_{2max}$ & $\lambda_{3max}$ & $\lambda_{4max}$ & $\lambda_{5max}$ & mIoU \%   \\ \hline
        0   & 481.00 & 481.00 & 481.00 & 481.00 & 481.00 & - \\ \hline
        50   & 481.78 & 482.40 & 480.63 & 481.46 & 482.12 & 10.20 \\ \hline
        100   & 481.60 & 482.53 & 479.68 & 481.12 & 481.76 & 13.17 \\ \hline
        150   & 481.37 & 482.48 & 478.96 & 480.74 & 481.40 & 17.67 \\ \hline
        200   & 481.73 & 482.96 & 478.98 & 480.98 & 481.69 & 21.26 \\ \hline
    \end{tabular}
\end{table*}

\subsubsection{Colour Vision for Hypothetical Martian Life}
\label{supple:evolution for mars}
In this section, we present predictions of possible colour vision capabilities for hypothetical Martian life, including dichromatic, trichromatic, and tetrachromatic vision.

Firstly, we train a 2-kernel opsin layer to optimize segmentation performance on the Mars-Seg dataset, aiming to predict the optimal dichromatic vision. As shown in Tab.\ref{tab:2 opsins on mars}, the initial wavelengths are set to values similar to human vision, 560 nm and 425 nm.

Secondly, we apply the gene duplication process to the previously trained 2-kernel opsin layer to analyse the evolutionary transition from dichromatic to trichromatic vision adapted to the Martian environment. Specifically, as shown in Tab.\ref{tab:3 opsins on mars}, the initial wavelengths are set to 588.81, 588.81, and 429.37 nm, corresponding to the optimal parameters obtained from the trained 2-kernel layer. It is worth noting that if the low-wavelength-sensitive opsin is allowed to duplicate, no duplication occurs, or more specifically, the two wavelengths remain indistinguishable and do not separate.

\begin{figure}[htbp]
  \centering

  \begin{minipage}[t]{0.43\textwidth}
    \centering
    \captionof{table}{Dichromatic Vision on Mars.}
    \label{tab:2 opsins on mars}
    \vspace{4pt}
       \begin{tabular}{c|cc|c}
        \hline
        Epoch & $\lambda_{1max}$ & $\lambda_{2max}$ &  mIoU    \\ \hline
        0   & 560.00 & 425.00 & - \\ \hline
        100   & 571.49 & 428.05 & 18.16 \\ \hline
        200   & 575.26 & 428.68 & 24.86 \\ \hline
        300   & 578.98 & 429.02 & 27.54 \\ \hline
        400   & 580.68 & 429.21 & 31.36 \\ \hline
        500   & 582.98 & 429.34 & 34.11 \\ \hline
        600   & 582.52 & 429.24 & 36.58 \\ \hline
        700   & 583.21 & 429.21 & 37.48 \\ \hline
        800   & 584.79 & 429.23 & 39.25 \\ \hline
        900   & 586.70 & 429.34 & 39.15 \\ \hline
        1000   & 588.81 & 429.37 & 41.53 \\ \hline
    \end{tabular}
  \end{minipage}
  \hfill
  \begin{minipage}[t]{0.51\textwidth}
    \centering
    \captionof{table}{Trichromatic Vision on Mars.}
    \label{tab:3 opsins on mars}
    \vspace{4pt}
      \begin{tabular}{c|ccc|c}
        \hline
        Epoch & $\lambda_{1max}$ & $\lambda_{2max}$ & $\lambda_{3max}$ & mIoU    \\ \hline
        0   & 588.81 & 588.81 & 429.37 & - \\ \hline
        100   & 596.48 & 584.76 & 431.02 & 16.44 \\ \hline
        200   & 603.89 & 580.89 & 432.35 & 23.79 \\ \hline
        300   & 609.40 & 576.38 & 433.12 & 26.22 \\ \hline
        400   & 615.76 & 573.13 & 434.03 & 29.77 \\ \hline
        500   & 621.05 & 573.67 & 434.30 & 32.71 \\ \hline
        600   & 625.98 & 574.24 & 434.57 & 35.30 \\ \hline
        700   & 630.70 & 575.09 & 434.83 & 36.38 \\ \hline
        800   & 637.33 & 577.62 & 435.22 & 37.77 \\ \hline
        900   & 642.24 & 579.63 & 435.55 & 39.14 \\ \hline
        1000   & 647.87 & 582.86 & 435.80 & 39.88 \\ \hline
    \end{tabular}
  \end{minipage}

\end{figure}

Similarly, we allow gene duplication to simulate the hypothetical evolutionary transition from trichromatic to tetrachromatic vision in the Martian environment, as illustrated in Tab.\ref{tab:4 opsins on mars}.

\begin{table}[htbp]
    \centering
    \caption{Tetrachromatic Vision on Mars.}
    \label{tab:4 opsins on mars}
    \begin{tabular}{c|cccc|c}
        \hline
        Epoch & $\lambda_{1max}$ & $\lambda_{2max}$ & $\lambda_{3max}$ & $\lambda_{4max}$ & mIoU    \\ \hline
        0   & 647.87 & 647.87 & 582.86 & 435.80 & - \\ \hline
        100   & 646.60 & 644.15 & 577.73 & 434.19 & 18.62 \\ \hline
        200   & 647.90 & 643.76 & 576.71 & 434.22 & 26.45 \\ \hline
        300   & 648.76 & 643.28 & 575.91 & 434.24 & 29.72 \\ \hline
        400   & 649.90 & 643.35 & 575.27 & 434.32 & 30.44 \\ \hline
        500   & 651.01 & 643.49 & 574.80 & 434.37 & 34.62 \\ \hline
        600   & 651.91 & 643.63 & 574.44 & 434.45 & 34.78 \\ \hline
        700   & 653.35 & 644.21 & 573.90 & 434.63 & 36.20 \\ \hline
        800   & 654.66 & 644.73 & 573.73 & 434.80 & 35.51 \\ \hline
        900   & 655.83 & 645.11 & 573.14 & 434.82 & 39.33 \\ \hline
        1000   & 657.08 & 645.72 & 572.63 & 434.95 & 41.07 \\ \hline
    \end{tabular}
\end{table}

The quantitative results indicate that dichromatic vision achieves the best performance in the Martian environment. This suggests that hypothetical Martian creatures may evolve to have dichromatic vision if the environmental conditions remain consistent.

\subsection{Supplementary Results for Camera Design}
\label{supple:camera}
\subsubsection{Camera for Martian Environment}
\label{supple:camera for mars}
In this section, we present the design of cameras with varying numbers of filters, optimized for operation in the Martian environment. As shown in Tab.\ref{tab:mars camera rb}, R, G, and B represent the red, green, and blue colour filters, respectively. The initial parameters for the maximum sensitivity wavelengths of these filters are set to 600 nm, 540 nm, and 460 nm, respectively, aligning with the values commonly used in general cameras \cite{camerafiltervalue}. Therefore, we train various configurations of the opsin layer with learning rate $3\times10^{-2}$ to obtain the colour response functions of filters optimized for the best segmentation performance. As shown in Tab.\ref{tab:mars camera rb}, we propose specific camera filter designs. For a minimalist Martian camera, the two-filter design with peaks at 624.22 and 459.67~nm offers a channel-count/performance trade-off: 40.78\% mIoU versus 41.60\% for four filters, a difference of 0.82 percentage points.

\begin{table}[htbp]
    \centering
    \caption{Camera Specialised for Martian Environment.}
    \label{tab:mars camera rb}
    \begin{tabular}{c|cccc|c}
        \hline
        Filters & $\lambda_{1max}$ & $\lambda_{2max}$ & $\lambda_{3max}$ & $\lambda_{4max}$ & mIoU \%   \\ \hline
        R   & 613.31 & - & - & - & 36.26 \\ \hline
        R/B   & 624.22 & 459.67 & - & - & 40.78 \\ \hline
        R/G   & 622.91 & 517.07 & - & - & 40.61 \\ \hline
        R/R/B   & 617.27 & 615.00 & 457.84 & - & 40.37 \\ \hline
        R/R/B/B   & 629.36 & 623.09 & 457.93 & 454.86 & 41.60 \\ \hline
    \end{tabular}
\end{table}

\subsubsection{Camera for Cancer Detection}
\label{supple:camera for cancer}
Similar to Sec.\ref{supple:camera for mars}, we train various configurations of the opsin layer to obtain the colour response functions of filters optimized for the best segmentation performance on MHSI dataset. As shown in Tab.\ref{tab:cancer camera rb}, we propose specific camera filter designs.

\begin{table}[htbp]
    \centering
    \caption{Camera Specialised for Cancer Detection.}
    \label{tab:cancer camera rb}
    \begin{tabular}{c|ccc|c}
        \hline
        Filters & $\lambda_{1max}$ & $\lambda_{2max}$ & $\lambda_{3max}$ & mIoU \%   \\ \hline
        R   & 612.95 & - & - & 54.22 \\ \hline
        R/B   & 620.43 & 477.91 & - & 56.97 \\ \hline
        R/G   & 623.44 & 538.26 & - & 56.64 \\ \hline
        R/R/B   & 621.11 & 619.56 & 472.21 & 54.53 \\ \hline
    \end{tabular}
\end{table}

In addition, for the segmentation task, we replace the MiT encoder and segmentation decoder used previously with a U-Net model \cite{unet} featuring a ResNet-34 backbone \cite{resnet} and a U-Net decoder, similar to the baseline model in \cite{cancerdetect}. The opsin layer is integrated into the U-Net architecture, and the opsin layer, backbone, and U-Net decoder are trained simultaneously. As illustrated in Tab.\ref{tab:cancer_unet}, we propose a 3-filter camera design optimized using the U-Net architecture, demonstrating that our evolutionary conservation optimization framework is robust and effective across different model architectures, including both transformer-based models with MiT encoders and convolutional architectures like U-Net.

 \begin{table}[htbp]
    \centering
    \caption{Train With U-Net Model.}
    \label{tab:cancer_unet}
    \begin{tabular}{c|ccc|c}
        \hline
        Camera & $\lambda_{1max}$ & $\lambda_{2max}$ & $\lambda_{3max}$ & Iou for cancer \%  \\ \hline
        Standard   & 600.00 & 540.00 & 460.00 & 52.33 \\ \hline
        Design   & 610.39 & 555.09 & 456.02 & 54.97 \\ \hline
    \end{tabular}
\end{table}

Due to the limited dataset size in our oral cancer detection task, we employ 5-fold cross-validation to robustly evaluate our method. To ensure a fair comparison between conventional RGB channels and our proposed adaptive camera design, we maintain identical experimental parameters (e.g., model architecture, learning rates) across both configurations. Each fold is repeated three times with randomized initializations, and we report the mean performance metrics to mitigate variance and enhance statistical reliability.

\subsection{Discussion, Limitation and Future Works}
\label{suppl:limitation}

In this paper, we present a paleoinspired vision framework that models retinal visual transduction and simulates evolutionary transitions of visual opsins. Our approach enables the quantitative analysis of opsin evolution with high computational efficiency, addressing a core challenge in traditional evolutionary biology. While methods like forward genetics experiments often suffer from resource-intensive stochastic searches for difficult-to-quantify traits \cite{Hawkins2021}, our framework provides algorithmic tools to streamline biological research. By implementing processes like evolutionary pressure analysis, we can quantify the direction and magnitude of trait changes in successive generations and simulate the specific environmental conditions of extinct species within an optimization-based paradigm.
Beyond theoretical insights, our framework establishes practical paradigms for high-performance camera design. The evolutionary conservatism embedded in our optimization supports manufacturing-friendly filter designs \cite{manufacturing}, while our minimalist approach, maximizing task-specific adaptability, aligns with emerging trends in computational imaging \cite{Minimalist_ECCV24}. 
Furthermore, the integration of direct photoreceptor control (Oz principle) suggests potential for engineering spectral sensitivities that bypass biological constraints to detect "impossible" colors \cite{oz-paper}. Such innovations could revolutionize surveillance, biomedical imaging, and the development of bio-compatible spectral encoding for brain-computer interfaces (BCIs), especially as the demand for vision-restoring neural chips continues to grow \cite{fda, braincomputerinterface}.

Despite advancing evolutionary insights, this study faces several constraints. First, the scarcity of functional or behavioral data limits the experimental validation of competing hypotheses, while relying on segmentation as a fitness proxy may underspecify the multifaceted nature of evolutionary pressures. Second, our environmental simulations and phylogenetic sampling remain simplified. Furthermore, the reliance on RGB-to-HSI reconstruction (via MST++), though state-of-the-art, is inherently ill-posed; the resulting spectral artifacts and ambiguities could introduce spurious correlations that confound evolutionary interpretations.

Future work will focus on several directions to address these limitations. On the evolutionary modeling front, we aim to enhance the framework by incorporating more sophisticated environmental and anatomical variables, and by collecting biologically relevant hyperspectral datasets to enable more rigorous paleontological analyses. On the camera design front, we plan to collaborate with materials scientists and optical engineers to prototype and validate the proposed camera designs, with careful consideration of manufacturing feasibility constraints including spectral bandwidth tolerances, fabrication precision, and quantum efficiency trade-offs. These efforts will bridge the gap between computational simulation and practical implementation, ultimately advancing both evolutionary biology research and task-specific imaging system development.


\clearpage
\subsection{Task-Based Evolutionary Modelling: Rationale and Evidence}
\label{cr:theoretical-discussion}
Our framework is intended as a computational hypothesis-testing model of visual adaptation, rather than a literal replay of evolutionary history. The central question is whether a biologically constrained visual phenotype---here, the number and spectral tuning of opsins---moves toward independently observed biological configurations when it is optimized under an explicit ecological demand. This section explains the origin of that modelling paradigm, why spatial recognition tasks are informative for the cases studied here, and what conclusions the resulting agreement can and cannot support.

\subsubsection{Origin in Computational Visual Ecology}
Using task performance to operationalize selection pressure predates this work. Morgenstern \textit{et al.} evolved biologically inspired networks against the statistics of natural spectra and obtained colour contrast and constancy without specifying them as separate objectives \cite{morgenstern2014natural}. Matthews \textit{et al.} combined measured fruit and leaf spectra with their spatial arrangement, then searched cone sensitivities that best preserved fruit--foliage patterns; the predicted M-cone optimum was close to the biological value \cite{matthews2018cone}. These studies establish a computational visual-ecology paradigm in which an ecological function is translated into a measurable visual objective and the resulting phenotype is compared with independent biological evidence.

Our work also follows the experimental logic of paleoinspired and robotics-inspired biology: a controllable surrogate is used to isolate a trait and evaluate its performance landscape when the corresponding historical system cannot be directly manipulated \cite{gravish2018robotics,paleoinspiredrobotics}. Our distinct contribution is to expose opsin number and peak sensitivity as the controlled phenotype, embed retinal spectral transduction in a trainable perception system, and test several documented transitions under matched environmental assumptions.

Subsequent independent work by Tiwary \textit{et al.} provides further support for this direction. Their embodied agents co-evolve eye morphology, neural processing, and behaviour under navigation, food--poison discrimination, and tracking objectives, and show that different task pressures lead to different visual systems \cite{tiwary2025whatifeye}. That study does not validate our individual reconstructions, but its results strengthen the broader premise that task-conditioned computational evolution can serve as a scientific sandbox for testing hypotheses about vision.

\subsubsection{Why Use a Semantic Task?}
We do not treat generic semantic segmentation accuracy as biological fitness. Instead, each task instantiates a specific survival-relevant visual demand. In MinneApple, apples are partially occluded by and spectrally confusable with foliage \cite{hani2019minneapple}. Recovering their category, location, boundary, and spatial extent therefore operationalizes fruit--foliage discrimination during foraging. Field observations provide an independent biological link: wild trichromatic capuchins consume conspicuously coloured fruit more efficiently than dichromats \cite{melin2017trichromacy}, while colour-vision phenotype also changes how capuchins combine visual and olfactory cues during food selection \cite{melin2019fruitscent}. Matthews \textit{et al.} further show that preserving the spatial pattern of fruit against leaves is informative for predicting cone tuning \cite{matthews2018cone}.

Segmentation is useful here because it jointly measures discrimination and figure--ground organization rather than only image-level recognition. Figure--ground separation is a fundamental operation in biological vision \cite{wagemans2012century,pasupathy2015neural}; in machine vision, promptable segmentation also transfers mask prediction across image distributions and downstream tasks \cite{kirillov2023segmentanything}. These observations motivate segmentation as a spatially informative interface, but they do not make mIoU a direct measurement of reproductive success. In our experiments, mIoU is an operational score for a controlled ecological task, and the learned spectral response is interpreted only after comparison with external biological observations.

\subsubsection{Distinguishing Adaptation from a Task Artifact}
A task-optimized spectrum could reflect the selected dataset or architecture rather than a general adaptive pressure. We therefore require converging evidence. First, the reconstructed direction must agree with measurements that were not used as optimization targets, such as documented opsin loss or duplication, spectral divergence, and depth-dependent RH1 blue shift. Second, the direction should persist when the downstream model changes. Third, related ecological objectives should produce comparable spectral solutions.

The experiments in Sec.~\ref{cr:additional-experiments} directly evaluate the latter two requirements. Across six segmentation encoders, the duplicated long-wave channels separate in the same direction. Segmentation, detection, and counting also recover closely located spectral optima while improving their respective task metrics. This does not prove that segmentation performance equals ecological fitness, but it substantially reduces the alternative explanation that the reported spectral adaptation is an artifact of one loss function or one backbone. The quantitative squirrelfish comparison provides a complementary test against specimen-based measurements rather than another machine-vision metric.

\subsubsection{Evidence Levels and Claim Boundary}
The biological cases do not have equal evidential status. Mammalian opsin loss and primate LWS/MWS duplication are strongly supported by molecular and comparative studies \cite{carvalho2017genetic,evolutionofverte1998,YOKOYAMA2000385}; the causal claim that frugivory was the primary pressure behind trichromacy is less certain \cite{advantageoftri}. Confidence is highest for the direction of squirrelfish RH1 blue shift because specimen measurements, phylogenetic reconstruction, and molecular analysis agree with the predicted trend \cite{squirrelfish}. The dim-light advantage of colour-vision deficiency remains uncertain because behavioural findings are limited, while multiple RH1 genes in deep-sea fish are well established but their specific causal relationship to bioluminescence is not directly proven \cite{multirod}. The Martian-life experiment is explicitly a speculative extrapolation with no biological validation.

Accordingly, agreement between optimized and measured spectra supports the framework as a controlled method for testing whether a proposed pressure is sufficient to produce a known direction of adaptation. It does not establish that the pressure was historically necessary, uniquely causal, or equivalent to organism-level fitness. Direct behavioural, genetic, phylogenetic, and physiological evidence remains the standard for those stronger claims.
\subsection{Additional Experimental Evidence}
\label{cr:additional-experiments}
This section evaluates the robustness of task-driven spectral adaptation across recognition architectures and objectives, and quantifies agreement with measured squirrelfish spectral sensitivities. General denotes the fixed initial responses and Design denotes the optimized responses within each comparison. These experiments test the dependence of the findings on the downstream model and task; they do not identify a unique historical evolutionary trajectory.

\subsubsection{Evolutionary Reconstruction Across Backbones}
\label{supple:task-backbone robustness}
The MinneApple experiments start from peaks of $(425,560,560)$~nm and compare six segmentation encoders. Design improves mIoU and reduces the camouflage score $S_R$ for every tested encoder (Table~\ref{tab:evolution backbone robustness}). The two longer-wavelength responses diverge in a similar direction, despite differences in the perception network.

The reported optimized short-wavelength peaks also differ from 425~nm. We therefore present the complete reported peak vectors without describing that channel as fixed in these additional experiments. The initialization is shared, but a fixed-channel constraint cannot be inferred from the table. The reported values are not accompanied by seed-level uncertainty estimates.

\begin{table*}[htbp]
    \centering
    \caption{Dichromacy-to-trichromacy reconstruction across segmentation encoders on MinneApple. Lower $S_R$ is better.}
    \label{tab:evolution backbone robustness}
    \resizebox{0.86\textwidth}{!}{
    \begin{tabular}{ll ccc cc}
        \toprule
        \textbf{Encoder} & \textbf{Opsin} &
        \textbf{$\lambda_{3\max}$} & \textbf{$\lambda_{2\max}$} & \textbf{$\lambda_{1\max}$} &
        \textbf{mIoU (\%)} & \textbf{$S_R$} \\
        & & \multicolumn{3}{c}{\textbf{(nm)}} & & \\
        \midrule
        MiT         & General & 425.00 & 560.00 & 560.00 & 47.29 & 0.276 \\
                    & Design  & 440.65 & 547.34 & 582.10 & 47.99 & 0.239 \\
        PVT-v2      & General & 425.00 & 560.00 & 560.00 & 49.21 & 0.265 \\
                    & Design  & 430.33 & 542.15 & 578.22 & 50.67 & 0.222 \\
        ConvNeXt    & General & 425.00 & 560.00 & 560.00 & 46.89 & 0.270 \\
                    & Design  & 434.42 & 545.93 & 584.41 & 47.52 & 0.239 \\
        SwinV2      & General & 425.00 & 560.00 & 560.00 & 48.66 & 0.255 \\
                    & Design  & 425.56 & 544.73 & 584.01 & 50.05 & 0.225 \\
        InternImage & General & 425.00 & 560.00 & 560.00 & 47.89 & 0.262 \\
                    & Design  & 420.10 & 533.26 & 576.12 & 48.65 & 0.238 \\
        MambaVision & General & 425.00 & 560.00 & 560.00 & 51.15 & 0.244 \\
                    & Design  & 426.64 & 545.03 & 579.91 & 52.21 & 0.219 \\
        \bottomrule
    \end{tabular}}
\end{table*}

\subsubsection{Reconstruction Across Recognition Tasks}
With the original MiT backbone, the additional experiments compare segmentation, detection, and fruit counting on MinneApple (Table~\ref{tab:cross task robustness}). Detection AP@50/AP@75 increases from 0.592/0.411 to 0.617/0.435, and counting MAE/RMSE decreases from 13.1/18.3 to 11.1/17.3. The learned longer-wavelength peaks remain similar across the three tasks. Thus, the observed direction of spectral adaptation is not confined to segmentation in these experiments. AP, F1, and counting errors retain their source units and are not converted into mIoU.

\begin{table*}[htbp]
    \centering
    \caption{Dichromacy-to-trichromacy reconstruction across downstream tasks on MinneApple.}
    \label{tab:cross task robustness}
    \resizebox{0.94\textwidth}{!}{
    \begin{tabular}{ll ccc l}
        \toprule
        \textbf{Task} & \textbf{Opsin} &
        \textbf{$\lambda_{3\max}$} & \textbf{$\lambda_{2\max}$} & \textbf{$\lambda_{1\max}$} &
        \textbf{Task-specific performance} \\
        & & \multicolumn{3}{c}{\textbf{(nm)}} & \\
        \midrule
        Segmentation & General & 425.00 & 560.00 & 560.00 & mIoU $=47.29\%$, $S_R=0.276$ \\
                     & Design  & 440.65 & 547.34 & 582.10 & mIoU $=47.99\%$, $S_R=0.239$ \\
        Detection    & General & 425.00 & 560.00 & 560.00 & AP@50/75 $=0.592/0.411$, mAP $=0.398$, F1 $=0.6805$ \\
                     & Design  & 439.22 & 545.24 & 581.45 & AP@50/75 $=0.617/0.435$, mAP $=0.409$, F1 $=0.7021$ \\
        Counting     & General & 425.00 & 560.00 & 560.00 & MAE/RMSE $=13.1/18.3$ \\
                     & Design  & 435.65 & 546.16 & 580.83 & MAE/RMSE $=11.1/17.3$ \\
        \bottomrule
    \end{tabular}}
\end{table*}

\subsubsection{Camera Design Across Backbones}
\label{supple:camera backbones}
On MODID and MHSI, Design improves mIoU over General in all twelve dataset--encoder combinations (Table~\ref{tab:camera backbone robustness}). Values in this table are fractions on $[0,1]$; for example, MiT on MODID improves from 0.5642 to 0.5867, equivalent to 2.25 percentage points. This supports complementarity between spectral optimization and backbone selection in the reported setting, without establishing significance in the absence of variances or confidence intervals.

\begin{table*}[htbp]
    \centering
    \caption{Camera-design robustness across segmentation encoders.}
    \label{tab:camera backbone robustness}
    \begin{tabular}{ll cc}
        \toprule
        \textbf{Encoder} & \textbf{Opsin} & \textbf{MODID mIoU} & \textbf{MHSI mIoU} \\
        \midrule
        MiT         & General & 0.5642 & 0.5435 \\
                    & Design  & 0.5867 & 0.5636 \\
        PVT-v2      & General & 0.5924 & 0.5670 \\
                    & Design  & 0.6057 & 0.5767 \\
        ConvNeXt    & General & 0.5350 & 0.5186 \\
                    & Design  & 0.5403 & 0.5250 \\
        SwinV2      & General & 0.5141 & 0.5539 \\
                    & Design  & 0.5426 & 0.5600 \\
        InternImage & General & 0.5725 & 0.5672 \\
                    & Design  & 0.5871 & 0.5812 \\
        MambaVision & General & 0.6155 & 0.5849 \\
                    & Design  & 0.6301 & 0.5946 \\
        \bottomrule
    \end{tabular}
\end{table*}

\subsubsection{Quantitative Agreement for Squirrelfish RH1}
\label{supple:blue shift agreement}
The learned RH1 peaks at depths of 0, 10, 50, and 70~m are 496.19, 488.73, 484.69, and 481.10~nm, respectively (Table~\ref{tab:blueshift}). Their strictly decreasing rank order gives Spearman $\rho=-1$, and their endpoint difference is 15.09~nm. The predicted range falls within the reported 481--502~nm range across nine squirrelfish species; the 70~m prediction differs by 0.10~nm from the reported 481~nm value for the deep-dwelling \textit{N.~aurolineatus}~\cite{squirrelfish}. This is an additional analysis of the existing depth experiment, not an independent training study or a species-matched estimate of historical causality.

\subsubsection{Interpreting the Supplementary Comparisons}
The full multi-rod trajectories in Tables~\ref{tab:single rod without}--\ref{tab:5 rods with} show a consistent interaction between channel count and simulated bioluminescence. At epoch 200, one/five rods achieve 11.14/13.77\% mIoU without bioluminescence and 15.17/21.26\% with it. The five-rod terminal peak spans are 0.33 and 3.98~nm, respectively. The larger separation supports a direction of adaptation within the model, while its small magnitude motivates richer ecological and photoreceptor models.

Filter-count experiments also expose task-dependent trade-offs. For MHSI, a single red filter yields 54.22\% mIoU and an R/B pair yields 56.97\%, an increase of 2.75 percentage points (Table~\ref{tab:cancer camera rb}). For Mars, two filters yield 40.78\% and four yield 41.60\% (Table~\ref{tab:mars camera rb}). These camera-design comparisons use camera-specific initializations and should not be pooled with the hypothetical Martian evolutionary trajectories.

The U-Net result in Table~\ref{tab:cancer_unet} is labelled cancer-class IoU, not mIoU. Its Standard/Design scores are 52.33/54.97\%. The source table does not identify a dataset and split unambiguously, so this result is retained as a separate architectural example rather than pooled with the MODID/MHSI backbone benchmark. Likewise, the stated five-fold, three-initialization protocol for oral cancer does not supply per-fold uncertainty and should not be interpreted as a significance test or automatically assigned to every additional experiment.
\clearpage


\newpage

\end{document}